\documentclass[preprint,12pt,authoryear]{elsarticle} 

\usepackage{amssymb}
\usepackage{amsmath}
\usepackage[nolist,nohyperlinks]{acronym}
\usepackage{amssymb}
\usepackage{amsthm}
\usepackage[labelfont=bf,justification=raggedright,singlelinecheck=false,font=footnotesize]{caption}
\usepackage{array}
\usepackage[utf8]{inputenc}
\usepackage[T1]{fontenc}
\usepackage{booktabs}
\usepackage{graphicx}
\usepackage{float}
\usepackage{lscape}
\usepackage{multirow}
\usepackage[nolist,nohyperlinks]{acronym}
\usepackage{caption}
\usepackage{subcaption}

\DeclareUnicodeCharacter{2212}{-}

\journal{Engineering Applications of Artificial Intelligence}

\begin{document}

\begin{frontmatter}

\title{Automatic Occupational Coding through Retrieval-Augmented Generation} 

\author[a,b]{Mathijs A. Langezaal}
\author[a]{Egon L. van den Broek\corref{cor1}}
\ead{e.l.vandenbroek@uu.nl}
\cortext[cor1]{Corresponding author}
\author[a]{Lukas P.A. Arts}
\author[b]{Marcel Goldberg}
\author[c]{Gr\'{e}goire Rey}
\author[d]{Melissa C. Friesen}
\author[d]{Sarah J. Locke}
\author[d]{Nathaniel Rothman}
\author[d]{Qing Lan}
\author[e]{Roel C.H. Vermeulen}
\author[e]{Susan Peters}

\affiliation[a]{organization={Department of Information and Computing Sciences, Utrecht University},
            addressline={Princetonplein 5},
            city={Utrecht},
            postcode={3584CC},
            country={The Netherlands}}
\affiliation[b]{organization={Population-Based Epidemiological Cohorts Unit UMS11, INSERM},
            addressline={16 Avenue Paul Vaillant Couturier},
            city={Villejuif},
            postcode={94807},
            country={France}}
\affiliation[c]{organization={France Cohortes UMS47, INSERM},
            country={France}}
\affiliation[d]{organization={Occupational and Environmental Epidemiology Branch, Division of Cancer Epidemiology and Genetics, National Cancer Institute},
            city={Bethesda},
            state={Maryland},
            country={USA}}
\affiliation[e]{organization={Institute for Risk Assessment Sciences, Utrecht University},
            addressline={Yalelaan 1},
            city={Utrecht},
            postcode={3584CL},
            country={The Netherlands}}

\begin{abstract}
Annually 1.5 million deaths and 76.1 million disability-adjusted life years are caused by exposure to agents during work (e.g., welding fumes, silica dust, and noise pollution). To improve worker health, high-quality exposure assessments are crucial. Therefore, free-text job descriptions are collected in large-scale occupational cohorts and manually standardized through occupational classifications. Subsequently, using Job Exposure Matrices (JEM), these classifications are linked to hazardous exposures. However, manual coding is slow, unreliable (i.e., 44--89\% inter-coder reliability), and costly, due to i) short, ambiguous free-text job descriptions, ii) classifications containing hundreds to thousands of categories, and iii) the absence of gold standards. With 46\%--79\% accuracy, state-of-the-art automatic occupational coding tools achieve classification performance on human level. However, their performance drops to 17--26\% when tested on external, out-of-distribution datasets. Furthermore, their supervised architectures limit them to occupational classifications present in the training set. Consequently, JEM utilizing other classifications are incompatible, leaving potentially vital exposure-response associations unexplored. With the Retrieval-Augmented Generation framework named OPERAG, we enable classification across any occupational classification without expert-labelled databases. OPERAG’s hybrid classification retrieves coding index information and combines this with Large Language Model-based classification. Validated on AsiaLymph’s 36,062 real-world job descriptions, OPERAG achieved 42.6\% occupational classification accuracy on the 4th coding level, substantially outperforming existing tools and approaching human coders' 51--56\% inter-coder reliability. Using two JEM named ALOHA+ and SHOCK, 89.9--98.6\% accuracy was achieved on exposure assessment. OPERAG approaches human-level, universal occupational classification. This enables efficient and effective large-scale exposure assessment, improving worker health.

\end{abstract}


\end{frontmatter}



\section{Introduction}
\noindent Work is a central part of adult life, with a large impact on overall health. Workers are exposed to various negative conditions, such as repetitive motions, noise pollution, and physical or chemical agents like welding fumes and silica dust~\citep{gbd2015global}. These occupational exposures are globally responsible for 1.5 million deaths annually and contribute 76.1 million disability-adjusted life years~\citep{GBD2016}. Epidemiological studies that assess exposure-response associations are crucial for fueling interventions to promote worker health, but they rely on high-quality exposure assessments to ensure correct and unbiased results~\citep{kriebel2007exposure}.

In large population-based cohorts, occupational data is typically collected through open-ended survey questions, yielding free-text job descriptions~\citep{hoffmann1995kind}. Subsequently, a Job Exposure Matrix (JEM) can be used to assign a wide range of exposures to occupations~\citep{peters2022occupational}. A JEM comprises estimated probabilities and intensities of exposure to harmful agents using standardized occupational titles and industries~\citep{peters2016syn}. Consequently, free-text job descriptions must first be standardized into such occupational groups using occupational classification systems. This process, known as occupational coding, is often performed manually by expert coders, which is time-consuming, expensive, and requires expert knowledge. Even with extensive documentation and training, consistent coding is challenging due to the large number of outcome categories in combination with the lack of a gold standard for each job description~\citep{hoffmann1995kind}. Consequently, many large-scale studies must deploy multiple expert coders, which introduces an inter-coder reliability ranging from 44\% to 89\%~\citep{mannetje2003use}.

To address manual occupational coding challenges, several automated coding tools have been developed with prediction accuracies ranging 46\%--79\% at the most detailed coding level~\citep{bao2020occupation, garcia2021s, gweon2017three, jones2004cascot, kouretsis2020labourr, langezaal2023artificial, park2015comparing, russ2016computer, savic2022procode, russ2023evaluation, russ2025}. These classification models have been implemented in Decision Support Systems (DSS) and semi-automatic coding tools that provide confidence scores (i.e., estimated correctness probability) alongside predictions~\citep{langezaal2023artificial, jones2004cascot, russ2016computer, russ2023evaluation}. This enables automatic coding of suggestions exceeding a customizable threshold, such as automatically coding all predictions with confidence scores above 95\%. At this 95\% threshold, estimated workload reductions range 19.7\%--55.7\%~\citep{langezaal2023artificial}. When predictions with confidence scores are presented to experts solely as suggestions without automatic coding (i.e., as a DSS), experts experience a 37.5\% workload reduction while maintaining outcome validity~\citep{langezaal2025operas}. 

\begin{figure}
    \centering
    \includegraphics[width=1\linewidth]{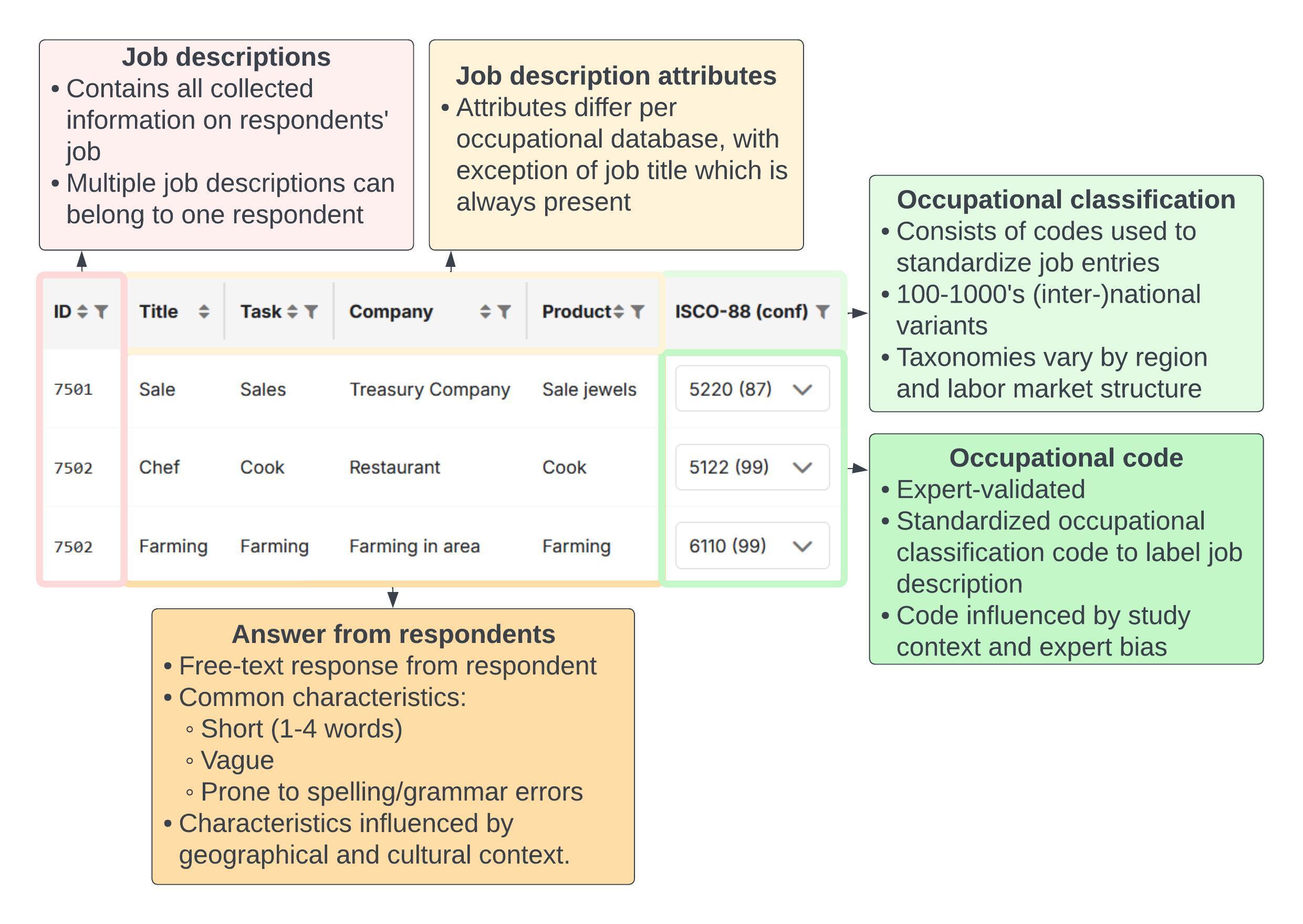}
    \caption{Overview of characteristics of job descriptions, their attributes, answers from respondents, occupational classifications, and occupational codes. Screenshot of the OPERAS decision support system~\citep{langezaal2025operas} showing three anonymized job entries from the AsiaLymph~\citep{friesen2016evaluation} dataset is used.}
    \label{fig:sceenshot}
\end{figure}

Current automatic coding tools predominantly use supervised ML approaches. These rely on training data from occupational databases, which contain job entries with accompanying labels, provided by experts. Both the entries and the labels possess specific characteristics (see Figure~\ref{fig:sceenshot}). The labels consist of attributes that have characteristics reflecting the geographical, cultural, and study context of the sample. The labels reflect the labor market from which the sample was drawn, as well as any study and expert biases. In the development of supervised ML models, the data is often divided into training and testing data. As the train and test samples are drawn from the same database and, hence, from the same data distribution, they have similar characteristics. Consequently, their performance on both the training and test data is similar~\citep{bao2020occupation, garcia2021s, gweon2017three, jones2004cascot, kouretsis2020labourr, langezaal2023artificial, park2015comparing, russ2016computer, savic2022procode, russ2023evaluation, russ2025}. When automatic coding models are applied on out-of-distribution (OoD) datasets (i.e., data sets with different job entry and label characteristics) instead of within-distribution datasets, their performance drops significantly, to 17\%–26\%~\citep{wan2023automated}. 

Publicly available expert-labeled occupational datasets remain scarce, which limits the development of supervised ML models with occupational classification coverage required for universal use in many different occupational epidemiological studies. To enable such universal use, alternative training sources are required, such as coding indices published alongside occupational classification systems~\citep{peterson1999occupational}. These indices provide labeled examples suitable for supervised training. However, these examples are limited (i.e., 1--5 per occupational group) and lack the heterogeneity of real-world free-text job descriptions. Consequently, models trained solely on these small, homogeneous corpora are prone to overfitting to coding index phrasing and will likely fail to generalize to real-world occupational datasets, limiting their OoD performance~\citep{chicco2017ten}. 

LLMs reduce the dependence on extensive expert-labeled datasets by utilizing their broad linguistic representations acquired during pretraining on massive corpora of textual data~\citep{naveed2025comprehensive}. Contrastingly to supervised ML models that may overfit to specific textual characteristics of occupational databases, LLMs can likely accurately process the linguistic variation inherent in free-text job descriptions, making them a promising alternative for AOC. However, despite these linguistic capabilities, LLMs' performance is often limited in data-scarce domains underrepresented in their pre-training corpora. When applied to such domains, these knowledge gaps frequently result in hallucinations where the model generates plausible-sounding but factually incorrect information, leading to increased classification errors. Furthermore, the plausible reasoning provided alongside a classification may mislead an expert into accepting an incorrect suggested code. Accurate classification of job descriptions is crucial for reliable exposure assessment and subsequent health outcomes. Consequently, the potential for such errors severely limits the reliable application of LLMs in data-scarce, high-stakes domains such as AOC.

To improve LLMs' performance in data-scarce domains, Retrieval‑augmented generation (RAG) supplements LLMs’ knowledge gaps with external, domain‑specific sources such as coding indices~\citep{lewis2020retrieval}. In addition to example job descriptions, these indices contain relevant information used by experts during occupational coding, including general descriptions and explanations for each occupational group~\citep{international2012international}. In addition to filling the LLMs' knowledge gap, this information can be used by LLMs to mimic expert-coder strategies to code occupations accurately, making RAG a promising approach for AOC. Hence, we developed OPERAG, a RAG framework enabling accurate OoD AOC without reliance on expert-labeled databases. We aim to address four specific AOC challenges: 1) the ambiguous and brief nature of job descriptions, 2) the high dimensionality of outcome categories, 3) the scarcity of public expert-labeled data, and 4) severe class imbalance. We address these challenges through the following contributions:
\begin{enumerate}
    \item Utilization of LLMs for AOC, harnessing their robust language understanding and reasoning capabilities to tackle AOC's ambiguous and short job descriptions and vast number of outcome categories.
    \item A processing pipeline that enables effective usage of occupational classifications' coding indices in RAG applications for AOC, eliminating the need for expert-labeled data.
    \item A RAG framework that uses coding indices as the sole knowledge base for AOC, ensuring the classification process remains independent of prior occupational distributions and thereby mitigating issues of class imbalance.
\end{enumerate}
\noindent The framework is assessed on occupational coding and exposure assessment performance through evaluation on a real-world occupational dataset and JEM applications. Consequently, this demonstrates the framework’s performance on occupational classification and its practical implications for exposure assessment in real-world occupational epidemiological studies.

The remainder of the paper is structured as follows: Section 2 introduces the conceptual background of occupational coding, as well as the technical background of LLMs and RAG. In Section 3, we describe the development of OPERAG, including the description of the evaluation dataset and ISCO‑88 coding index used to enable AOC without expert-labeled databases. Additionally, we detail the construction of the RAG component, including its structure‑aware knowledge base and retriever, followed by the reasoning‑LLM reranking component, LLM-classifier design, and inference protocol. Subsequently, we outline the evaluation methodology for both occupational coding and exposure assessment. In Section 4, we present OPERAG's evaluation results, including benchmarks against expert coders and other automatic coding tools. We conclude in Section 5 with a discussion of practical implications, limitations, and avenues for future work.

\section{Background}
\subsection{Occupational coding}
\noindent In large-scale occupational cohorts, occupational information is collected from respondents in free-text format~\citep{hoffmann1995kind}. Such job descriptions are often ambiguous, incomplete, contain misspellings, and are frequently very short (i.e., 2-4 words per description). To standardize these free-text job descriptions, occupational classification systems are employed~\citep{international2012international}. These systems contain hundreds to thousands of hierarchically structured occupational groups, where each group is denoted by multiple characters (e.g., 3433). The first character denotes the most aggregated occupational group, with each subsequent character adding detail until the most specific group is reached, denoted by the last character. This information is often published in the coding index alongside the occupational classification. The coding index contains the classification's structure, each group's definitions, and example job descriptions for most occupational groups. Although sharing similarities in structure, occupational classification systems differ across countries to reflect differences in national labor markets and social contexts~\citep{emmel20102010}. Consequently, hundreds of national and international occupational classification systems have been developed, each following its own taxonomy of occupational groups.

Manual occupational coding is typically performed by expert coders, which is time-consuming, expensive, and requires specialized knowledge~\citep{burstyn2014beyond}. Even after extensive training and experience, coders generally process only approximately 2,700 job episodes per month. This inherently slow and variable nature of the occupational coding process stems from occupational descriptions lacking a single, objective "gold-standard" code, except for the specific examples in the coding index~\citep{international2012international}. Consequently, expert coders must rely on a combination of the coding index's definitions and supplementary coding rules. These rules can be generally applicable (e.g., inclusion or exclusion criteria specified in the index) and context-dependent rules (e.g., how to handle incomplete or ambiguous occupational information or cases where only coding to a more aggregated coding level is possible)~\citep{hoffmann1995kind}. Such rules are crucial for ensuring consistent and accurate occupational coding to secure reliable results in the following study. However, although considered the gold-standard for occupational coding, manual coding's inter-coder reliability only ranges from 42\% to 71\%~\citep{maaz2009intercoder, russ2016computer, massing2019standardized, conrad2016classifying}. Beyond inter-coder variability, expert coders are also susceptible to intra-coder variability, as a single coder may assign different codes to the same occupational information at different coding sessions~\citep{patel2012performance}. Combined, these factors highlight the inherent variability and difficulty of occupational coding.

To address the challenges of manual occupational coding, several automatic coding tools have been developed for different languages and occupational classification systems. These tools enable automatic classification of job descriptions into occupational classifications without the need for human intervention. They employ different classification architectures, ranging from word-matching techniques to LLM-based classifiers. Consequently, classification performance differs, ranging 46\%--79\% on the most detailed coding level when tested on a subset obtained from the model's training dataset (i.e., the model's within-distribution performance)~\citep{bao2020occupation, garcia2021s, gweon2017three, jones2004cascot, kouretsis2020labourr, langezaal2023artificial, park2015comparing, russ2016computer, savic2022procode, russ2023evaluation, russ2025}. However, performance on data not stemming from their training dataset is much lower due to the large difference in data characteristics between occupational datasets, including description length and used terminology. For example, the automatic coding systems developed by~\citep{kouretsis2020labourr, jones2004cascot, garcia2021s} achieved an OoD classification performance of 17\%, 26\%, and 20\%, respectively. This shows current tool's inability to be effectively used in occupational studies using datasets other than the datasets used during model training. An overview of current automatic coding tools, used occupational classifications, and within-distribution classification performance can be found in Table~\ref{tab:auto_tools}.

\begin{table}[H]
\centering
\caption{Overview of currently available automatic coding systems, with their year of release, supported classification system(s) (CS), and performance. The coding level (CL), number of outcome categories (OC), and within-distribution Accuracy (Acc, \%) are specified for each CS' least aggregated coding level.}
\label{tab:auto_tools}
\renewcommand{\arraystretch}{1.15}
\setlength{\tabcolsep}{6pt}
\begin{tabular}{l l l c r r}
\hline
\textbf{Author(s)} & \textbf{Year} & \textbf{CS*} & \textbf{CL} & \textbf{OC} & \textbf{Acc.}\\
\hline
\multirow{1}{*}{Bao et al. \nocite{bao2020occupation}} & \multirow{1}{*}{2020} & Canadian NOC2016 & 4 & 500 & 59\%\\
\hline
\multirow{1}{*}{Garcia et al. \nocite{garcia2021s}} & \multirow{1}{*}{2021} & Canadian NOC2016 & 4 & 500 & 65\%\\
\hline
\multirow{1}{*}{Gweon et al. \nocite{gweon2017three}} & \multirow{1}{*}{2017} & ISCO\text{-}08 & 4 & 412 & 65\%\\
\hline
\multirow{3}{*}{Jones et al. \nocite{jones2004cascot}} & \multirow{3}{*}{2004} & ISCO\text{-}08 & 4 & 412 & 64\%\\
 &  & UK SOC (2010, 2000, 90) & -- & -- & --\\
 &  & UK SIC (2007, 2003, 92, 80) & -- & -- & --\\
\hline
\multirow{1}{*}{Kouretsis et al. \nocite{kouretsis2020labourr}} & \multirow{1}{*}{2020} & ISCO\text{-}08 & 4 & 436 & --\\
\hline
\multirow{4}{*}{Langezaal et al. \nocite{langezaal2023artificial}} & \multirow{4}{*}{2023} & PCS2003 & 4 & 497 & 69\%\\
 &  & NAF2008 & 5 & 732 & 79\%\\
 &  & ISCO\text{-}88 & 4 & 390 & 61\%\\
 &  & ISCO\text{-}68 & 5 & 1506 & 58\%\\
\hline
\multirow{2}{*}{Park et al. \nocite{park2015comparing}} & \multirow{2}{*}{2015} & US SOC2010 & 6 & 840 & 49\%\\
 &  & US NAICS2012 & 6 & 270 & 51\%\\
\hline
\multirow{1}{*}{Russ et al. \nocite{russ2025}} & \multirow{1}{*}{2025} & US NAICS 2022 & 6 & 1012 & 47\%\\
\hline
\multirow{2}{*}{Savic et al.\nocite{savic2022procode}} & \multirow{2}{*}{2022} & PCS2003 & 4 & 497 & 57\%\\
 &  & NAF2008 & 5 & 732 & 63\%\\
\hline
\multirow{1}{*}{Russ et al.\nocite{russ2023evaluation}} & \multirow{1}{*}{2023} & US SOC2010 & 6 & 840 & 50\%\\
\hline
\multirow{3}{*}{Russ et al.\nocite{russ2025soccernet}} & \multirow{3}{*}{2025} & US SOC1980 & 4 & 664 & 69\%\\
 &  & Canadian NOC2011 & 4 & 500 & 74\%\\
 &  & ISCO\text{-}88 & 4 & 390 & 46\%\\
\hline
\multirow{1}{*}{Russ et al.\nocite{russ2016computer}} & \multirow{1}{*}{2016} & US SOC2010 & 6 & 840 & 51\%\\
\hline
\end{tabular}
\caption*{*Classification system abbreviations: ISCO = International Standard Classification of Occupations; NOC = National Occupational Classification; SOC = Standard Occupational Classification; SIC = Standard Industrial Classification; PCS = Professions et Catégories Socioprofessionnelles; NAF = Nomenclature d'Activités Française; NAICS = North American Industry Classification System.}
\end{table}

Only a limited number of occupational databases are publicly available, with most usually containing only a single occupational classification. For example, the U.S. O*NET database provides detailed information on occupational groups aligned with the Standard Occupational Classification (SOC) system~\citep{handel2016net}, while the European ESCO framework maps occupations to ISCO-08 categories~\citep{icociu2019esco}. Consequently, for occupational classifications that lack publicly available expert-coded databases, the coding index remains the only publicly accessible resource.

\subsection{Large Language Models}
\noindent Deep Learning (DL) architectures, such as feedforward and long short-term memory networks, have demonstrated state-of-the-art performance across diverse classification tasks, including text-classification~\citep{minaee2021deep}. By employing multiple layers of interconnected neurons, these models automatically derive representations of data with multiple levels of abstraction, thereby minimizing the dependence on manually engineered features typically required for traditional ML approaches. However, their high parameterization requires extensive datasets to achieve competitive performance~\citep{liu2017deep}. Consequently, in domains with limited example-outcome pairs or homogeneous datasets that fail to capture the heterogeneity of the target domain, DL approaches are prone to overfitting or underfitting. Consequently, this restricts their generalizability and limits their successful application in such data-scarce environments~\citep{neyshabur2017exploring}.

LLMs reduce DL’s dependence on large labeled datasets through transformer-based architectures and large-scale unsupervised pretraining on diverse text corpora, resulting in base models with extensive linguistic representations and contextual understanding \citep{naveed2025comprehensive}. After pretraining, these base models typically undergo additional supervised fine-tuning to improve conversational fluency, instruction adherence, or to adapt their internal knowledge and linguistic properties to specific domains \citep{peng2023instruction}. These include instruction-tuned LLMs, which are trained on datasets containing human-aligned instruction–response pairs, demonstrating strong dialogue and task-following abilities. However, LLMs' performance is limited in domains that are insufficiently represented in their pretraining data~\citep{yang2024harnessing}. To address this limitation, domain-specific fine-tuning adjusts an LLM’s parameters to capture domain terminology and knowledge~\citep{anisuzzaman2024fine}. However, the performance of such models is limited for complex tasks~\citep{ge2023openagi}. Here, performance improves when "Chain-of-Thought" prompting is applied, which is a technique that explicitly prompts the model to produce a series of intermediate reasoning steps before generating the final output. To further improve performance in complex tasks, LLMs have been post-trained on curated datasets containing step-by-step reasoning-answer pairs~\citep{kumar2025llm}. Through this training, models generate structured intermediate reasoning steps during inference, achieving state-of-the-art performance across a wide range of tasks, including text classification.

Interaction with an LLM typically occurs through a user-submitted prompt that specifies the query, after which the model generates a corresponding response~\citep{marvin2023prompt}. Such interactions can be preceded by a predefined system prompt to establish the model's overall behavior or tone. Both prompt types can be engineered using various prompting structures and strategies (i.e., prompt engineering) to optimize performance for specific tasks. For text classification applications, the system prompt commonly instructs the LLM to assume the role of an expert within the relevant classification domain~\citep{clavie2023large}. Subsequently, the user prompt is structured following a consistent format, comprising: 1) a task description, 2) the input data requiring classification, and 3) an output schema to enable consistent parsing. To guide the model's reasoning through domain-relevant classification steps, task-specific classification rules can be specified in the user-prompt~\citep{mu2023can}. Additionally, prompting strategies such as chain‑of‑thought (CoT) prompting are often employed to further encourage step-by-step reasoning before final label prediction~\citep{yeo2025demystifying}. Through the combination of these approaches, the LLM may mimic an expert's decision-making process, which has been shown to yield state-of-the-art classification performance for task-specific applications.

A major constraint of LLMs is their fixed knowledge after pre-training, determined by the contents and cutoff date of their pretraining corpus~\citep{patil2024review}. Consequently, performance in data-scarce or highly specialized domains is often limited, often resulting in responses with hallucinations~\citep{feng2024don}. These are linguistically coherent but factually incorrect or irrelevant responses. When such outputs are misinterpreted as accurate, they pose significant risks, especially in high-stakes contexts related to health. To overcome these limitations, the aforementioned domain fine-tuning can result in more accurate responses~\citep{peng2023instruction}. However, this process still requires large datasets containing relevant, heterogeneous example-output pairs to yield robust, generalizable model performance~\citep{rahman2025large}. Therefore, frameworks that enable effective domain-specific specialization in data-scarce settings are crucial for improving LLM reliability and accuracy in niche domains.

\subsection{Retrieval-Augmented Generation}
\noindent Retrieval-Augmented Generation (RAG) is a framework designed to mitigate the limitations of LLMs' knowledge gaps by integrating external knowledge sources into the generation process~\citep{lewis2020retrieval}. Here, RAG supplements the LLMs' knowledge with dynamic, query-specific knowledge retrieved from validated sources such as domain-specific databases, curated corpora, or structured knowledge graphs. This enables LLMs to adapt to underrepresented domains and data shifts without model retraining or finetuning~\citep{yu2024evaluation}. 

A typical RAG system comprises two components, namely 1) a retriever, which identifies relevant information from a knowledge base, and 2) a generator, which returns an output based on the retrieved content and user's query~\citep{lewis2020retrieval}. The generator, usually an LLM, uses the retriever's retrieved content in combination with the user's query to generate an output. Although the choice of generator has a high impact on subsequent classification performance, RAG's ability to effectively retrieve and supply the generator with relevant knowledge hinges on the setup and development of its retriever~\citep{zhao2024retrieval}, consisting of three components:
\begin{itemize}
    \item \textbf{Knowledge Base}, which is the set of external information from which evidence can be retrieved;
    \item \textbf{Indexing Process}, which is the pre-processing procedure that prepares information for retrieval;
    \item \textbf{Retrieval Function}, which is the algorithm used for selecting, ranking, and returning a list of relevant entries
\end{itemize}
\noindent Subsequently, as the retrieval function is generally optimized for efficiency and broad recall, a more accurate and computationally expensive reranker may subsequently be applied to produce a shorter, more accurate candidate list~\citep{zhang2025qwen3}. In the following paragraphs, each component will be explained in more detail.

The knowledge base contains all retrievable relevant information~\citep{su2025parametric}. It can include documents such as research papers, technical manuals, or news articles, as well as structured resources like ontologies, knowledge graphs, or databases. To ensure all LLMs' knowledge gaps relevant to the current tasks are covered, it is essential that relevant, task-specific information is included. 

The indexing process adapts, converts, and stores task-specific information to enable accurate retrieval from its knowledge base~\citep{sakar2025maximizing}. Typically, this process consists of three stages: chunking, embedding, and vector storage. Chunking divides large documents into smaller, semantically coherent units to enhance retrieval performance. Here, when semantic boundaries are unclear, fixed-length splits can be used~\citep{stabler2025impact}. However, this may result in scattering related information across different chunks, hindering effective retrieval. Contrastingly, semantic or structure-aware segmentation detects contextual shifts or uses inherent document structure to maintain semantic coherence within chunks~\citep{qu2024semantic}. After chunking, embedders map text chunks into high-dimensional vectors where semantically similar content are near to each other in the embedding space, allowing similarity search functions to identify the most relevant passages to the query~\citep{akbik2019flair}. Consequently, retrieval performance depends on the embedder’s ability to assign similar embeddings to queries and relevant content, for which several types of embedders are available~\citep{darrin2024embedding}. These include general-purpose embedders that capture broad meaning, domain-specific ones that improve performance for domain-terminology, and instruction-aware embedders that integrate explicit task instructions into embeddings, ensuring results are not only semantically similar but also aligned with the retrieval’s intended purpose.

The retrieval function is used to determine which chunks of information are most relevant to the user's query~\citep{cuconasu2024power}. Here, similarity metrics are often employed, including cosine similarity, inner product, or L2 distance. These metrics calculate what information chunks are closest to the query in the embedding space, thereby aiming to retrieve the most relevant information. However, such similarity-based metrics often fail to capture deeper contextual, logical, or task-specific relationships between the query and the indexed information~\citep{gan2024similarity}. Consequently, the top‑k candidates retrieved may include irrelevant or only partially related results, which can degrade performance due to LLMs’ position bias~\citep{shi2024judging}. This bias is a systematic tendency to weigh information differently depending on its position in the input sequence rather than purely on its semantic relevance, which can manifest in several ways, namely:
\begin{itemize}
    \item over‑attending to the beginning or end of a prompt regardless of their importance;
    \item under‑utilizing information in the middle of long contexts (the “lost in the middle” phenomenon);
    \item favoring items purely based on list order (e.g., the first option) or formatting position when comparing alternatives.
\end{itemize}
\noindent These effects limit accurate reasoning and classification, particularly in long‑context tasks and multi‑item comparisons such as occupational coding~\citep{zhang2024can}. 

To ensure the most relevant entries from the retrieval function are in the top positions when provided to the generator, rerankers are often employed~\citep{zhang2025qwen3}. These modules assign relevance scores by jointly evaluating the query and each candidate passage~\citep{li2024can}. In cross‑encoder-based rerankers, the query–candidate pair is jointly encoded to produce an independent relevance score for each candidate, after which candidates are sorted accordingly~\citep{li2024can}. Because the model attends across both texts token by token, this approach often yields higher retrieval accuracy than purely semantic similarity search. However, since cross‑encoders primarily assess semantic relationships, their usefulness is limited in tasks requiring rule-based or multi‑step reasoning~\citep{dejean2024thorough}. Contrastingly, LLM-based rerankers can incorporate multi‑step reasoning and rule-based evaluation, leveraging their advanced linguistic and contextual understanding to identify complex query–document relationships~\citep{ren2025self}. Moreover, they enable the integration of domain-specific rules and logic into the reranking process, improving retrieval accuracy in ambiguous or reasoning-intensive scenarios.

\section{Methods}
\begin{figure}
    \centering
    \includegraphics[width=1\linewidth]{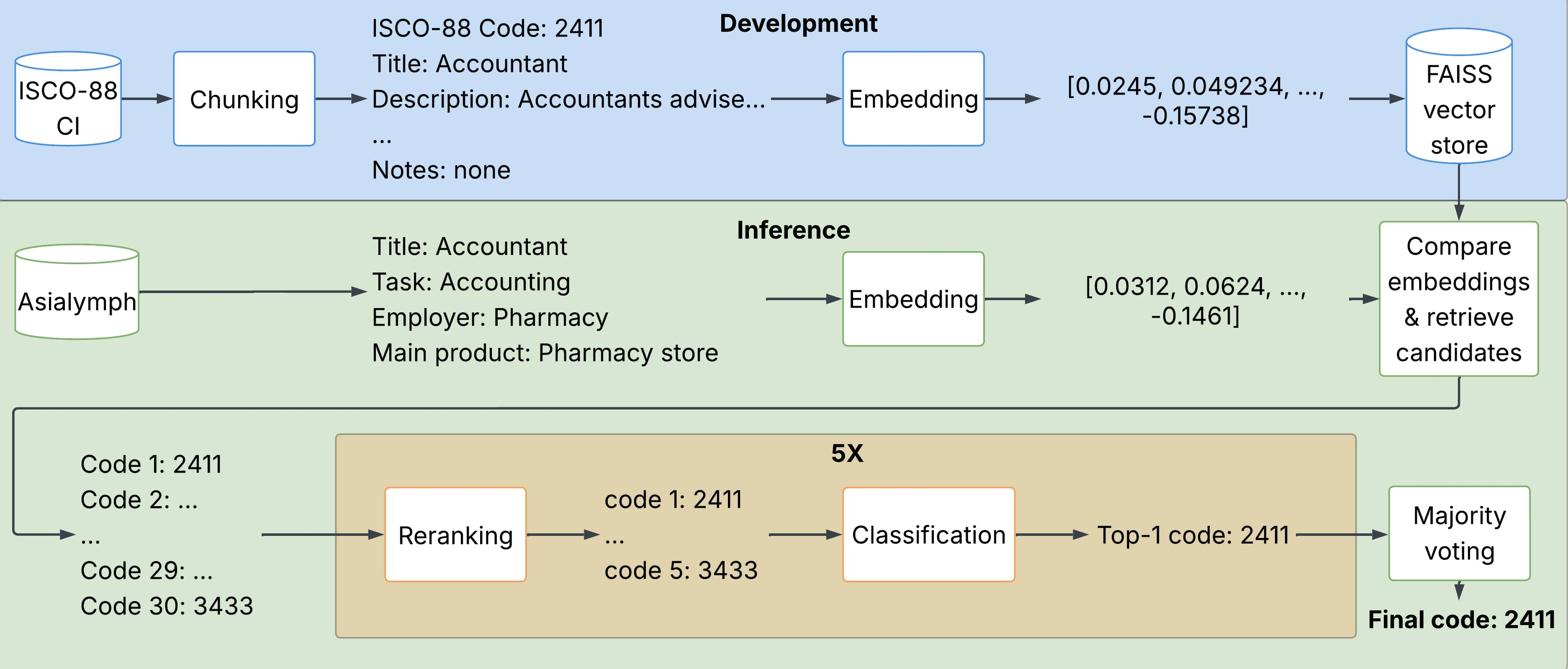}
    \caption{Overview of OPERAG's retriever component development and overall classification pipelines, illustrated by the example job description "Accountant". The development phase is indicated in blue, while the inference process during the classification phase is shown in green. Orange color indicates components that are executed iteratively.}
    \label{fig:OPERAG_dev}
\end{figure}

\noindent We developed the OPERAG framework in five stages: First, we described the evaluation dataset, its data characteristics, and employed occupational classification. Second, we developed OPERAG's retriever through identifying key knowledge-base information, in this case, the occupational classification's coding index, and used its document structure to chunk the information into occupation group chunks for effective retrieval. Here, we identified an effective storage and retrieval mechanism. Third, we implement a reasoning-LLM-based reranker module to rerank the top-k retrieved candidates using expert-coder-based classification rules. Fourth, we developed OPERAG's generator, which employs the same reasoning-LLM and classification rules as the reranker module. Here, we designed a classification prompt using multiple prompt-engineering techniques and inference protocol for optimal OoD classification performance. Fifth, we detail the performance evaluation for both occupational coding and exposure assessment using commonly employed metrics. An overview of the OPERAG system can be found in Figure~\ref{fig:OPERAG_dev}. In the following sections, each stage will be explained in detail.

\subsection{Dataset}
\noindent To evaluate OPERAG’s OoD classification performance, we used the AsiaLymph dataset~\citep{friesen2016evaluation}. This dataset contains real-world English occupational information coded with ISCO‑88, allowing a real-world occupational coding evaluation. AsiaLymph is a prototypical occupational dataset due to its short-text data characteristics and its use of the widely applied ISCO-88 classification (see Table \ref{tab:desc_stat}). Furthermore, AsiaLymph has been widely employed for training and benchmarking job coding tools, particularly for assessing OoD performance, thereby enabling direct and fair comparison of classification results. In addition, AsiaLymph's coding protocol and inter-coder reliability have been reported, providing a reliable basis for comparison against human coders. 

AsiaLymph is a hospital‑based case–control study of lymphoma among Chinese populations in Eastern Asia~\citep{friesen2016evaluation}. It includes free-text job episodes in traditional and simplified Chinese containing job titles, tasks, employers, and employers' main products. This information was manually coded into the 5th revision of the Standard Occupational Classification System of the Republic of China (SOCS-ROC)~\citep{DGBAS1992}. Subsequently, the information was expert-translated to English, with the codes crosswalked to ISCO-88. Two study centers independently coded the occupational data in the original language, with discordances and crosswalk discrepancies (i.e., in the case of a one-to-many mapping) being resolved by a third expert coder. In our evaluation, we use the job episodes' English translation with corresponding ISCO-88 codes.

ISCO‑88 comprises 544 outcome categories (i.e., occupational groups) in four aggregation levels, each represented by a digit (e.g., “4521”): 10 major groups (e.g., “4”), 28 sub‑major groups (e.g., “45”), 116 minor groups (e.g., “452”), and 390 unit groups (e.g., “4521”)~\citep{peterson1999occupational}. Here, each additional digit adds specificity to the coded occupational information. When the occupational information lacked sufficient detail to code to the most specific level, trailing zeros were used to denote non-codable levels (e.g., “4500”). In case an episode cannot be coded to any occupational group, this was denoted by "zzzz". The descriptive statistics of AsiaLymph can be found in Table~\ref{tab:desc_stat}.  

\begin{table}[H]
\caption{\textbf{Descriptive statistics of the AsiaLymph dataset} 
(IQR = interquartile range).}
\label{tab:desc_stat}
\centering
\begin{tabular}{@{}l r@{}}
\toprule
\textbf{Statistic} & \textbf{Count} \\
\midrule
Number of job episodes & 36,179 \\
Number of unique outcome categories & 389 \\
Number of input classes & 4 \\
Minimum word count & 1 \\
Maximum word count & 52 \\
Median word count (IQR) & 10 (5) \\
Minimum outcome category size & 1 \\
Maximum outcome category size & 2,251 \\
Median category size (IQR) & 26 (69) \\
\bottomrule
\end{tabular}
\end{table}

\subsection{Retriever}
\noindent The development of OPERAG's retriever consisted of three stages: 1) knowledge-base identification and construction, 2) Identifying the optimal chunking strategy, embedder, and vector database for effective coding index information storage, and 3) developing the retrieval mechanism, including prompt-design for effective retrieval using the instruction-aware embedder, and identifying the optimal similarity metric. In the following paragraphs, each stage will be explained in more detail.

Since OPERAG is exclusively used for occupational classification using ISCO-88, OPERAG's knowledge base consists solely of ISCO-88 coding index's relevant information. The ISCO-88 coding index contains the detailed information for each occupational group, including the group title, definition, included tasks, example occupations, similar occupations classified elsewhere, and notes that guide occupational coding~\citep{peterson1999occupational}. As experts often use all aforementioned information during occupational coding, we ensured OPERAG's knowledge base contained all this information.

As the ISCO-88 coding index is organized into discrete occupational group entries, we leveraged this structure to apply the structure-aware segmentation chunking strategy to divide the occupational group's information into semantically coherent units for optimal retrieval performance. Consequently, this resulted in 523 chunks, each containing all the aforementioned information of one occupational group. An example chunk is provided in Appendix A. Subsequently, we embedded each chunk into dense numerical vectors encoding their semantic meaning using the instruction-aware Qwen3‑Embedding‑8B~\citep{zhang2025qwen3}. We selected this embedder for its state-of-the-art retrieval performance across multiple retrieval benchmarks, enabling efficient and accurate similarity search against the to be classified occupational information. Finally, we stored these vectors in a FAISS vector database, which is developed to enable accurate similarity search and retrieval over the knowledge base~\citep{douze2025faiss}.

The retrieval mechanism contains two phases: 1) embedding of the to be classified job description, including instructions to leverage the instruction-aware capabilities of the employed embedder, and 2) comparing this embedding against the embeddings in the knowledge base using a similarity metric. Consequently, we first aggregate all available information of a job episode into a query in combination with the following instruction: "Given an occupation's Title, Task(s), Employer, and Employer's main product, retrieve relevant passages that match the occupation". By subsequently embedding this using ``Qwen3-Embedding-8B'' employed during the indexing process, we ensure consistency in the semantic representation across both the query and knowledge base entries while leveraging the embedder's instruction-aware retrieval capabilities for optimal retrieval performance~\citep{zhang2025qwen3}. Once embedded, the query vector is compared to all stored occupational group vectors in the knowledge base using the L2 (Euclidean) distance metric. This approach enables us to effectively quantify semantic similarity and identify entries that are most closely aligned with the to-be-coded occupational information. Here, we retrieve the top 30 coding index entries with the smallest L2 distances to the query embedding, representing the entries most relevant to the occupational information. This information is subsequently used in the reranker module.

\subsection{Reranker}
\noindent As classification performance in downstream tasks generally degrades with increasing numbers of tokens in an LLM’s prompt, it is beneficial to reduce the number of coding index entries included in the classification LLM’s context~\citep{an2024make}. Furthermore, LLMs exhibit the aforementioned position bias, which is the systematic tendency to weigh information differently depending on its position in the input sequence rather than purely on its relevance to the query~\citep{shi2024judging}. Consequently, to limit LLM's context length and position bias problems, we introduce a reranking module. Here, an auxiliary model re‑evaluates the candidate occupational codes retrieved in the first stage and reordering them based on relevance to the occupational information~\citep{li2024can}. This ensures that the most probable codes are prioritized, while less relevant candidates are demoted or excluded. Here, we select a top-5 higher-quality subset from the top-30 retrieved codes, ordering them based on relevance to the to-be-coded job description before classification.

Purely similarity‑based reranking using conventional cross-encoder rerankers is insufficient for reliable separation of highly similar occupational codes~\citep{dejean2024thorough}. This is due to occupational groups in the same detailed group often being semantically similar, therefore occupying near‑identical regions of the embedding space~\citep{langezaal2023artificial}. Hence, explicit application of coding index rules (e.g., differentiation by required skill level) is often required for accurate separation of these codes and, subsequently, accurate occupational coding~\citep{hoffmann1995kind}. For example, ISCO‑88 codes 3433 (Bookkeepers) and 4121 (Accounting and Bookkeeping Clerks) list highly similar tasks but are performed at different skill levels~\citep{peterson1999occupational}. However, this information is often not explicitly stated in individual job descriptions, which warrants the application of coding rules to determine the appropriate skill level and, consequently, the correct code. This may require considering prior occupations or other contextual information (e.g., if a person has never performed accounting tasks, they are more likely to be a bookkeeper than an accountant, despite a job title stating "accountant"). Therefore, a reranker that takes into account such coding rules is required for accurate reranking of occupational codes.

To allow for the application of coding rules and other relevant context in the reranking process, we employ a reasoning-LLM (i.e., Qwen3‑14B-AWQ) as the reranker module~\citep{yang2025qwen3}. This model is prompted to select and order the top-30 codes from the retrieval stage into the top-5 most likely codes based on the provided occupational information. The reasoning is guided by rules mimicking expert-coder coding, using the following prompt structure:
\begin{itemize}
    \item \textbf{System Role:} Indicate that the model is acting as an expert-coder, trained to apply the coding index rules consistently;
    \item \textbf{Context:} The top-30 retrieved ISCO‑88 coding index entries;
    \item \textbf{Objective:} Instruction to propose the top-5 most likely ISCO‑88 codes for the given occupation in order of relevance;
    \item \textbf{Classification Rules:} Expert coder mimicking coding rules to guide the reranking;
    \item \textbf{Input Data:} The occupational information to classify;
    \item \textbf{Output Format:} Standardized schema to return the top-5 predicted ISCO‑88 codes in order of relevance, comma-separated, and followed by a concise reasoning section.
\end{itemize}
\noindent The reranking stage's full prompt can be found in Appendix B.

To maximize reranking performance, we used the following recommended hyperparameter settings during inference: \textit{temperature} = 0.6, $top_p$ = 0.95, $min_p$ = 0, $top_k$ = 20, and \textit{presence penalty} = 1.5~\citep{yang2025qwen3}. To reduce stochastic variability inherent to LLMs (i.e., the same occupational information can elicit different top-5 occupational code lists when prompted multiple times), each case is reranked five times separately with identical prompts and settings~\citep{chen2024more}. Subsequently, each ordered list is passed to a separate classifier in the classification stage to identify the most relevant occupational code to the occupational information.

\subsection{Classification LLM}
\noindent As mentioned, occupational coding is a highly rule‑intensive task that requires strong reasoning capabilities to disambiguate between the often very similar occupational groups~\citep{kumar2025llm}. Therefore, we employ a reasoning-LLM to select the top-1 occupational code for the to be classified occupational information from the top-5 codes from the reranking module.

Since occupational data are often proprietary as they contain patient or participant data, these datasets cannot be transmitted to third‑party providers for inference. However, most leading LLMs are either (1) proprietary, requiring external transmission of data, or (2) open‑source but so computationally demanding (e.g., Deepseek R1) that they cannot be realistically deployed on consumer‑grade hardware available to individual expert-coders and/or study centers~\citep{guo2025deepseek}. Consequently, we employ a compact, reasoning-LLM, Qwen3‑14B‑AWQ, to classify occupations into the most suitable occupational code using the top‑5 candidates obtained from the reranking stage~\citep{yang2025qwen3}. Qwen3‑14B‑AWQ has demonstrated state-of-the-art reasoning and instruction-following performance for its parameter count (i.e., 14 billion), while remaining computationally efficient enough to run on consumer‑grade hardware. This allows accurate automatic occupational classification to be performed locally, enabling accurate, efficient, and data privacy-preserving AOC.

To mimic expert‑coder coding practice and ensure consistent performance, each classification prompt follows a fixed, structured template including expert-coder rules to maximize classification performance, similar to the prompt used during the reranking stage~\citep{clavie2023large}. Additionally, to further improve performance, the model is instructed to reason step‑by‑step (i.e., Chain‑of‑Thought reasoning) to determine which coding index entry best fits the occupational information~\citep{chen2025towards}. Finally, the output must follow a standardized schema consisting of: (1) the predicted code at the most specific permissible level and (2) a brief justification showing decisive inclusion/exclusion criteria. This schema simplifies parsing and validation, and supports explainability, which can be vital for trust in AI predictions, as often required for decision-support systems~\citep{jung2024trust}. The classification stage's full prompt can be found in Appendix C.

As five different top-5 most relevant codes from the reranking stage are provided to the classifier five separate times, the final prediction is selected by majority vote. This limits the stochastic variability inherent to LLMs, while improving classification performance~\citep{chen2024more}. Here, let $\mathcal{C} = \{c_1, c_2, \dots, c_n\}$ be the set of $n=5$ classifications for a single job episode. The final prediction, $\hat{c}_{final}$, is the code that appears most frequently in $\mathcal{C}$:

\begin{equation}
    \hat{c}_{final} = \underset{c' \in \mathcal{U}}{\text{arg max}} \sum_{i=1}^{n} \mathbb{I}(c_i = c'),
\end{equation}

where $\mathcal{U}$ is the set of unique codes present in $\mathcal{C}$, and $\mathbb{I}(\cdot)$ is the indicator function, which is 1 if the condition is true and 0 otherwise. In the event of a tie, the system re‑prompts iteratively until a majority is obtained. If an output violates the schema (e.g., multiple codes, missing fields, or unsupported formats), the system automatically re‑prompts until a single occupational code is obtained. For each job episode, this final occupational code is subsequently used during the performance evaluation.

\subsection{Performance evaluation}
\noindent To enable an extensive evaluation of OPERAG's OoD performance, we use two widely used evaluation metrics employed in similar studies, namely: Accuracy and PABAK. This enables a direct OoD comparison with humans and other tools employing the same metrics~\citep{wan2023automated}. We assess OPERAG's performance on occupational classification and exposure assessment accuracy. In the following sections, we explain each evaluation stage in more detail.

\subsubsection{Occupational classification performance}
\noindent Using the AsiaLymph dataset, we assessed percent-agreement and inter-coder reliability using Accuracy. For each job episode, we consider the manually expert-coded ISCO-88 code from AsiaLymph as the gold standard.

We deem accuracy to be an important metric in the current study, as it indicates the proportion of occupational codes that will potentially require less coding time during the occupational coding process. We define accuracy as the proportion of correctly predicted codes relative to the total number of predictions~\citep{demsar2006statistical, witten2002data}. Given a confusion matrix over $K$ categories, where the diagonal elements $n_{ii}$ represent the number of correctly classified episodes for each category $i$, and $N$ is the total number of episodes, accuracy is calculated as:

\begin{equation}
    \text{Accuracy} = \frac{\sum_{i=1}^{K} n_{ii}}{N}.
\end{equation}

\noindent For each classification model, we computed accuracy per code level, the most aggregated occupational group, and the three most frequent 4-digit codes for each most aggregated occupational group.

As occupational datasets exhibit highly imbalanced label distributions due to differences in the prevalence of occupations performed across populations, we deem the F1-score to be an important metric in this study. The F1-score provides a harmonic mean of precision and recall, thereby accounting for both false positives and false negatives in the classification process~\citep{yacouby2020probabilistic}. This is particularly relevant when some occupational groups are substantially more frequent than others, as the F1-score prevents the model from achieving high performance simply by predicting the majority classes~\citep{pivin-bachler2025simba}. Additionally, this metric enables reliable performance estimation for individual categories while accounting for the distribution within the dataset. For a given occupational category, the F1-score is defined as:

\begin{equation}
\text{F1} = 2 \times \frac{\text{Precision} \times \text{Recall}}{\text{Precision} + \text{Recall}},
\end{equation}

\noindent where Precision represents the proportion of correctly predicted episodes within a category relative to all episodes predicted for that category, and Recall represents the proportion of correctly predicted episodes relative to all episodes that truly belong to that category. We computed the macro-averaged F1-score across all occupational categories to provide equal weight to each category regardless of its frequency in the dataset. Additionally, we calculated the F1-score for each major group to account for the imbalance in occupations performed in the AsiaLymph dataset.

\subsubsection{Exposure assessment evaluation}
\noindent To enable a comprehensive exposure assessment evaluation over a multitude of different exposures, we assessed OPERAG's exposure assessment performance using the ALOHA+-JEM and SHOCK-JEM~\citep{matheson2005biological},~\citep{huss2013electric}. In addition to computing Accuracy using Eq. (2), we also employ PABAK, a metric commonly used in occupational studies for assessing inter-coder reliability~\citep{mak2004prevalence}. This is due to the aforementioned imbalanced label distribution in occupational datasets, which PABAK explicitly accounts for. PABAK does not use empirical marginals to compute chance agreement and is therefore particularly suitable for instances where label distribution is highly imbalanced. This enables us to 1) obtain a reliable estimation of exposure assessment performance not influenced by label distribution, and 2) facilitate comparison against other occupational studies. Following~\citet{wan2023automated}, we dichotomized each exposure into unexposed (level 0) and exposed (levels 1 and 2) before computing both metrics, enabling a direct comparison with the automatic coding tools previously evaluated on the same dataset. PABAK is defined as:

\begin{equation}
    \text{PABAK} = 2P_o - 1,
\end{equation}

\noindent where $P_o$ is the observed proportion of agreement between the dichotomized exposure assignments based on the gold-standard and the OPERAG-predicted codes~\citep{mak2004prevalence}.

The ALOHA+-JEM has originally been developed for a study assessing the association of occupational exposure and symptoms of chronic bronchitis and pulmonary ventilatory defects in a general population-based study of five areas in Spain~\citep{matheson2005biological}. Subjects completed a respiratory questionnaire on symptoms and occupation and underwent baseline spirometry. Based on expert knowledge, the JEM was developed ad hoc by two occupational experts. It contains a job axis and an exposure estimate for 12 generic industrial exposures, namely: biological dust, mineral dust, gas fumes, VGDF (i.e., vapors, gases, dust, and fumes), all pesticides, herbicides, insecticides, fungicides, aromatic solvents, chlorinated solvents, other types of solvents, and metals. Exposure estimates for each aforementioned exposure can be obtained for all ISCO-88 codes. Here, three levels of exposure can be assigned, namely 0: no exposure, 1: low exposure, and 2: high exposure. For each gold standard manually coded and OPERAG-predicted ISCO-88 code, we applied the ALOHA+-JEM to obtain a level of exposure for each aforementioned exposure, and subsequently calculated percent-agreement and PABAK for each exposure separately. In line with~\citet{wan2023automated}, codes without an exposure estimate in the JEM (i.e., codes only assignable at an aggregated level, such as 6000) were assigned exposure level 0.

The SHOCK-JEM was developed to estimate the occupational risk of electric shocks using routinely collected national accident statistics from five European countries (Germany, UK, Austria, Switzerland, and the Netherlands)~\citep{huss2013electric}. Electrical injury cases were drawn from national registries, and denominators (i.e., workers per occupation) were obtained from EUROSTAT, enabling calculation of annual electrical injury rates per 10,000 workers by occupation and country, which were then pooled with a random-effects model to account for heterogeneity across countries. The resulting SHOCK-JEM provides an exposure estimate for each 3rd level ISCO-88, with 0 being low, 1 being medium, and 2 being high chance of experiencing electric shocks. We applied the SHOCK-JEM to the gold standard 3-digit manually, and OPERAG predicted ISCO-88 codes to obtain the exposure assessment for all job episodes. Here, we computed overall accuracy and PABAK. Similar to the ALOHA+-JEM, we dichotomized the exposure estimate into low (0) versus medium or high (1--2) chance.

To account for the chance level of a non-exposure assignment being substantially higher, given the many occupational codes in the JEMs resulting in no exposure, we evaluated two groups: 1) all individuals and 2) exposed individuals. For the first group, we calculated the aforementioned metrics using all individuals in the AsiaLymph dataset. For the second group, we only included job episodes in the evaluation that were deemed exposed according to the gold-standard expert codes. For each exposure, we considered an episode exposed if its gold-standard exposure level for that exposure exceeded 0. Furthermore, assigning an exposed individual a not-exposed status often has the largest effect on the following epidemiological study. Hence, evaluating this group will give more insight into the applicability of these classification models in real-world scenarios and allow for a direct comparison with OPERAS' ML-based classification models. Here, we compute accuracy and PABAK for both groups.

\section{Results}
\noindent After data cleaning, 36,062 entries remained for the performance evaluation. 117 job episodes could not be evaluated because AsiaLymph did not contain a gold-standard code for these episodes (i.e., episodes coded as "zzzz"). We report OPERAG's overall system performance on both occupational classification and exposure assessment below. Additionally, since OPERAG's RAG and reranker components have separate influences on occupational classification performance, we also report performance metrics for this evaluation of each component individually. An overview of the classification performance and the exposure assessment evaluation using the ALOHA+-JEM can be found in Tables \ref{tab:major_group_metrics_comparison} and \ref{tab:aloha_jem}, respectively.

\subsection{Classification performance evaluation}
\begin{figure}
    \centering
    \includegraphics[width=1\linewidth]{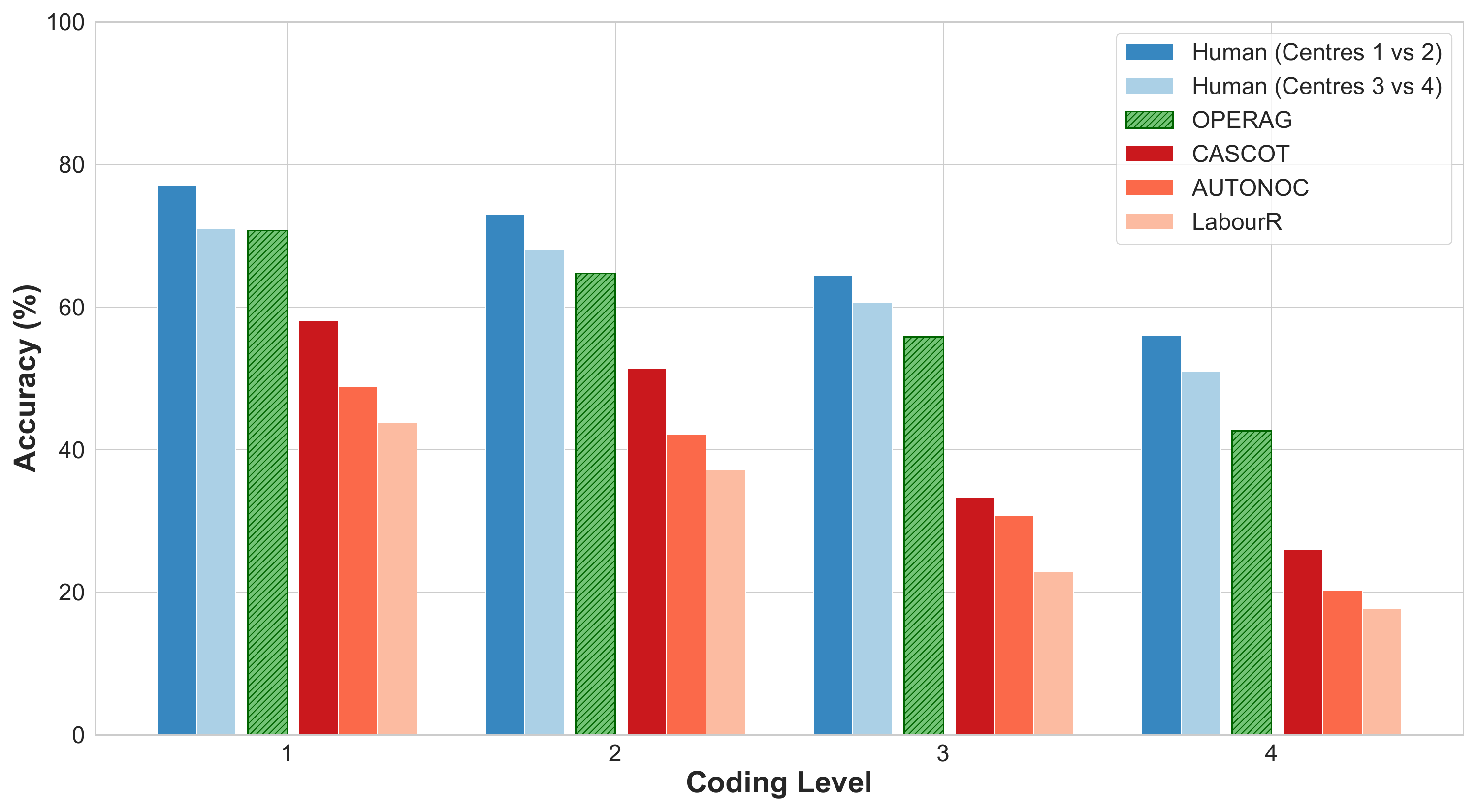}
    \caption{Occupational classification Accuracy (\%) for all of ISCO‑88's coding levels (1–4) for human-coders, OPERAG, and automated coding tools (AUTONOC, CASCOT, LabourR) on the AsiaLymph dataset. “Centres 1 vs 2” and “Centres 3 vs 4” denote inter‑coder agreement between different coding centres during the AsiaLymph coding process, as reported in~\citep{ge2021occupational}.}
    \label{fig:class_comparison}
\end{figure}

\noindent OPERAG's overall system achieved state-of-the-art OoD automatic classification performance compared to established occupational coding tools (see Figure~\ref{fig:class_comparison}). Specifically, OPERAG reached accuracies of 70.7\% (F1 = 0.70), 64.7\% (F1 = 0.50), 55.8\% (F1 = 0.46), and 42.6\% (F1 = 0.39) on ISCO‑88’s 1st, 2nd, 3rd, and 4th coding levels, respectively. Contrastingly, existing automatic coding tools demonstrated considerably lower OoD performance on the same AsiaLymph dataset. Here, AUTONOC, CASCOT, and LabourR achieved only 43.8–58.1\% at the 1st level, 37.2–51.4\% at the 2nd level, 22.9–33.3\% at the 3rd level, and 17.7–26.0\% at the 4th level of ISCO‑88, with CASCOT consistently performing best across all coding levels.

Performance across major occupational groups ranged from 13.6\% for Legislators, Senior Officials, and Managers to 84.4\% for Armed Forces, with Professionals achieving the highest accuracy among the remaining major groups (69.8\%). In comparison, AUTONOC and LabourR showed substantially lower performance, ranging 6.6–46.4\% and 4.5–37.7\%, respectively, with the same groups being their worst and best performers. CASCOT had slightly higher performance, with accuracies ranging from 9.6–54.5\%. However, its lowest performing major group was Skilled Agricultural and Fishery Workers (9.6\%). Here, OPERAG achieved 23.3\% accuracy for this group.

\begin{table}[H]
\centering
\caption{Classification performance on AsiaLymph (n = 36,062) by ISCO-88 major group (MG). The \% column shows the frequency distribution of each MG in AsiaLymph. OPERAG results include both F1 score and accuracy (Acc, \%), while AUTONOC, CASCOT, and LabourR report Acc. only. N/A indicates unreported metrics.}
\label{tab:major_group_metrics_comparison}
\begin{tabular}{@{}crrrrrr@{}}
\toprule
& & \multicolumn{2}{c}{\textbf{OPERAG}} & \textbf{AUTONOC} & \textbf{CASCOT} & \textbf{LabourR} \\
\cmidrule(lr){3-4}
\textbf{\shortstack{MG*}} & \textbf{\%} & \textbf{F1} & \textbf{Acc.(\%)} & \textbf{Acc.(\%)} & \textbf{Acc.(\%)} & \textbf{Acc.(\%)} \\
\midrule
1 & 4.5  & 0.31 & 13.6 & 6.6  & 11.6 & 4.5 \\
2 & 7.1  & 0.58 & 69.8 & 42.9 & 46.0 & 30.4 \\
3 & 15.0 & 0.48 & 27.8 & 12.7 & 16.0 & 12.9 \\
4 & 9.3  & 0.38 & 38.0 & 11.2 & 17.8 & 15.5 \\
5 & 13.0 & 0.62 & 62.3 & 46.4 & 54.5 & 37.7 \\
6 & 12.9 & 0.38 & 23.3 & 10.8 & 9.6  & 6.0 \\
7 & 11.9 & 0.45 & 45.8 & 17.3 & 27.8 & 15.5 \\
8 & 16.7 & 0.45 & 45.2 & 15.8 & 27.9 & 19.9 \\
9 & 7.4  & 0.46 & 46.7 & 20.7 & 22.7 & 13.2 \\
0 & 2.1  & 0.92 & 84.4 & N/A  & N/A  & N/A  \\
\bottomrule
\end{tabular}
\caption*{
*Major group numbers correspond to ISCO-88 major group classifications: 1, Legislators, senior officials, and managers; 2, Professionals; 3, Technicians and associate professionals; 4, Clerks; 5, Service workers and shop and market sales workers; 6, Skilled agricultural and fishery workers; 7, Craft and related trades workers; 8, Plant and machine operators and assemblers; 9, Elementary occupations; 0, Armed Forces.}
\end{table}

\begin{table}[H]
\centering
\caption{OPERAG's exposure assessment evaluation performance of the ALOHA+-JEM. Accuracy (Acc., \%) and PABAK are given for all individuals and exposed individuals. Here, N represents the number of exposed individuals for this exposure. Exposure was dichotomized into unexposed (0) and exposed (1--2).}
\label{tab:aloha_jem}
\begin{tabular}{@{}lrrrrr@{}}
\toprule
\multirow{2}{*}{\textbf{Exposure Type}} & \multicolumn{2}{c}{\textbf{All individuals}} & \multicolumn{3}{c}{\textbf{Exposed individuals}} \\
\cmidrule(lr){2-3} \cmidrule(lr){4-6}
& \textbf{Acc.} & \textbf{PABAK} & \textbf{N} & \textbf{Acc.} & \textbf{PABAK} \\
\midrule
Biological dust & 91.7 & 0.83 & 12,062 & 88.5 & 0.77 \\
Mineral dust & 91.7 & 0.83 & 14,588 & 90.1 & 0.80 \\
Gas/fumes & 90.2 & 0.80 & 19,621 & 91.4 & 0.83 \\
VGDF* & 92.8 & 0.86 & 22,093 & 94.3 & 0.89 \\
All pesticides & 96.9 & 0.94 & 5,270 & 90.1 & 0.80 \\
Herbicides & 98.6 & 0.97 & 4,203 & 97.7 & 0.95 \\
Insecticides & 98.0 & 0.96 & 4,575 & 96.3 & 0.93 \\
Fungicides & 97.4 & 0.95 & 4,867 & 91.2 & 0.82 \\
Aromatic solvents & 89.9 & 0.80 & 8,635 & 72.7 & 0.45 \\
Chlorinated solvents & 93.9 & 0.88 & 4,046 & 76.2 & 0.52 \\
Other solvents & 91.6 & 0.83 & 6,478 & 75.1 & 0.50 \\
Metals & 95.2 & 0.90 & 4,305 & 80.9 & 0.62 \\
\bottomrule
\end{tabular}
\caption*{*VGDF refers to the exposures: vapors, gases, dust, and fumes.}
\end{table}

OPERAG's accuracy of the three most prevalent ISCO-88 codes of each major occupational group in the AsiaLymph dataset varied widely: 2.3–54.6\% for Legislators, Senior Officials and Managers; 83.5–95.1\% for Professionals; 1.5–44.6\% for Technicians and Associate Professionals; 4.8–44.7\% for Clerks; 57.0–60.7\% for Service Workers and Shop and Market Sales Workers; 3.9–84.1\% for Skilled Agricultural Workers; 30.2–80.7\% for Craft and Related Workers; 24.4–44.0\% for Plant and Machine Operators and Assemblers; and 21.3–65.0\% for Elementary Occupations. A detailed overview of classification performance for these top three codes within each major group, along with the three most frequently predicted codes corresponding to them, is provided in Table~\ref{tab:isco_with_titles}.

\begin{table}[H]
\centering
\caption{Top incorrect predictions for the three most prevalent ISCO-88 unit groups (UG) per major group (MG) in AsiaLymph. For each UG, its frequency in AsiaLymph (\%, total \textit{n} = 36,062), accuracy, and the top three most frequent incorrect predictions with their share of predictions (\%) within their MG are given.}
\label{tab:isco_with_titles}
\resizebox{\textwidth}{!}{%
\begin{tabular}{@{}ll>{\raggedright\arraybackslash}p{8cm}cccc@{}}
\toprule
\textbf{MG*} & \textbf{UG} & \textbf{Occupational group title} & \textbf{Acc.} & \multicolumn{3}{c}{\textbf{Top incorrect predictions}} \\
\cmidrule(l){5-7}
& & & \textbf{(\%)} & \textbf{1st} & \textbf{2nd} & \textbf{3rd} \\
\midrule
0 & 0110 (2.1\%) & Armed forces & 84.4\% & 3132 (1.5\%) & 5169 (1.2\%) & 1229 (0.8\%) \\
\midrule
1 & 1210 (2.6\%) & Directors and Chief Executives & 2.3\% & 1314 (21.8\%) & 1319 (14.4\%) & 1312 (11.4\%) \\
& 1224 (0.3\%) & Production and Operations Department Managers in Wholesale and Retail Trade & 11.3\% & 1314 (30.4\%) & 1315 (14.8\%) & 1225 (8.7\%) \\
& 1222 (0.2\%) & Production and Operations Department Managers in Manufacturing & 54.6\% & 1312 (15.9\%) & 1232 (3.4\%) & 1210 (2.3\%) \\
\midrule
2 & 2331 (1.1\%) & Primary Education Teaching Professionals & 95.1\% & 2359 (1.3\%) & 1229 (1.0\%) & 3310 (0.8\%) \\
& 2320 (0.8\%) & Secondary Education Teaching Professionals & 91.5\% & 1229 (2.6\%) & 2300 (2.0\%) & 2331 (1.6\%) \\
& 2221 (0.8\%) & Medical Doctors & 83.5\% & 3221 (4.8\%) & 3241 (2.2\%) & 3231 (1.8\%) \\
\midrule
3 & 3439 (4.9\%) & Administrative Associate Professionals not classified elsewhere & 1.5\% & 1222 (16.5\%) & 1229 (10.4\%) & 1231 (8.7\%) \\
& 3433 (1.7\%) & Bookkeepers & 2.3\% & 4121 (55.5\%) & 2411 (20.2\%) & 1231 (7.3\%) \\
& 3415 (1.5\%) & Technical and Commercial Sales Representatives & 44.6\% & 5220 (23.7\%) & 3420 (4.0\%) & 3429 (2.7\%) \\
\midrule
4 & 4100 (2.2\%) & Office Clerks & 44.7\% & 3430 (11.3\%) & 4000 (6.5\%) & 4190 (5.6\%) \\
& 4111 (1.6\%) & Stenographers and typists & 4.8\% & 4100 (42.8\%) & 4190 (13.6\%) & 4000 (4.1\%) \\
& 4131 (1.4\%) & Stock Clerks & 40.3\% & 1235 (11.7\%) & 9333 (6.2\%) & 1226 (3.3\%) \\
\midrule
5 & 5220 (5.2\%) & Shop Salespersons and Demonstrators & 57.0\% & 5230 (10.6\%) & 3415 (7.6\%) & 1314 (2.8\%) \\
& 5122 (2.1\%) & Cooks & 60.7\% & 5120 (21.7\%) & 7412 (4.5\%) & 9111 (2.6\%) \\
& 5123 (1.8\%) & Waiters, Waitresses and Bartenders & 57.5\% & 5120 (25.0\%) & 9132 (3.3\%) & 5220 (1.9\%) \\
\midrule
6 & 6110 (7.1\%) & Market gardeners and crop growers & 3.9\% & 6111 (47.3\%) & 9211 (16.6\%) & 6130 (14.5\%) \\
& 6111 (2.2\%) & Field crop and vegetable growers & 67.2\% & 9211 (20.7\%) & 6130 (2.8\%) & 6110 (2.0\%) \\
& 6130 (0.8\%) & Market-oriented crop and animal producers & 84.1\% & 9211 (3.1\%) & 9210 (3.1\%) & 6111 (1.7\%) \\
\midrule
7 & 7233 (1.1\%) & Agricultural- or industrial-machinery mechanics and fitters & 30.2\% & 7230 (41.7\%) & 7241 (5.1\%) & 7240 (4.9\%) \\
& 7212 (0.7\%) & Welders and flame cutters & 80.7\% & 8283 (5.0\%) & 7213 (1.9\%) & 7210 (1.9\%) \\
& 7241 (0.7\%) & Electrical mechanics and fitters & 37.3\% & 7240 (14.5\%) & 7230 (8.3\%) & 7137 (6.6\%) \\
\midrule
8 & 8322 (1.7\%) & Car, taxi and van drivers & 43.2\% & 8320 (29.2\%) & 9151 (6.3\%) & 0110 (4.1\%) \\
& 8211 (1.5\%) & Machine-tool operators & 44.0\% & 8210 (10.1\%) & 7222 (3.5\%) & 7224 (3.2\%) \\
& 8263 (1.3\%) & Sewing-machine operators & 24.4\% & 7436 (30.1\%) & 7430 (30.1\%) & 8260 (2.3\%) \\
\midrule
9 & 9322 (1.7\%) & Hand packers and other manufacturing labourers & 47.3\% & 9320 (18.0\%) & 9333 (4.7\%) & 9300 (4.7\%) \\
& 9132 (1.3\%) & Helpers and cleaners & 65.0\% & 9130 (6.0\%) & 9141 (4.8\%) & 5120 (4.0\%) \\
& 9312 (1.1\%) & Construction and maintenance labourers & 21.3\% & 9313 (26.3\%) & 9300 (17.0\%) & 9310 (16.8\%) \\
\bottomrule
\end{tabular}%
}
\caption*{
*Major group numbers correspond to ISCO-88 major group classifications: 1, Legislators, senior officials, and managers; 2, Professionals; 3, Technicians and associate professionals; 4, Clerks; 5, Service workers and shop and market sales workers; 6, Skilled agricultural and fishery workers; 7, Craft and related trades workers; 8, Plant and machine operators and assemblers; 9, Elementary occupations; 0, Armed Forces.}
\end{table}

The RAG component achieved a top-30 retrieval accuracy of 87.8\% and a top-5 accuracy of 65.1\%. The correct code was retrieved in the 1st, 2nd, 3rd, 4th, and 5th positions in 34.5\%, 11.6\%, 7.4\%, 6.0\%, and 5.6\% of cases, respectively. After applying the reranking stage, the top-5 accuracy increased to 69.4\%, with the correct code now appearing in the 1st, 2nd, 3rd, 4th, and 5th positions in 41.4\%, 14.4\%, 6.6\%, 4.2\%, and 2.6\% of cases, respectively.

\subsection{Exposure assessment evaluation}
\noindent The median exposure assessment accuracy and PABAK of all ALOHA+-JEM's exposures were 93.3\% (range: 89.9\%–98.6\%), and 0.87 (range: 0.80–0.97) for all individuals, respectively. Similarly, OPERAG achieved state-of-the-art exposure assessment performance when compared against the coding tools AUTONOC, CASCOT, and LabourR. These tools had a lower median PABAK of 0.69 (range: 0.46--0.79), 0.78 (range: 0.59--0.94), and 0.70 (range: 0.42--0.94), respectively, with OPERAG outperforming all tools on all exposures reported in the ALOHA+JEM. Similar results were found for the SHOCK-JEM, where OPERAG had an accuracy and PABAK of 92.0\% and 0.84, respectively. Compared, AUTONOC, CASCOT, and LabourR had a lower median PABAK on the SHOCK-JEM of 0.56, 0.67, and 0.49, respectively.

OPERAG's exposure evaluation using the ALOHA+JEM for exposed individuals only resulted in a median accuracy and PABAK of 90.1\% (range: 72.7\%–97.7\%), and 0.80 (range: 0.45–0.95), respectively. For evaluation with the SHOCK-JEM, this was 83.7\% and 0.67 for the same metrics, respectively. This discrepancy between exposure assessment performance for all individuals and exposed individuals warrants careful attention when applying AOC to ensure accurate and stable results of the following occupational epidemiological study.

\section{Discussion}
\noindent OPERAG achieved state‑of‑the‑art out‑of‑distribution performance in both occupational classification and exposure assessment, substantially outperforming other automatic coding tools while approaching human‑level accuracy on the same dataset. Here, occupational classification accuracy on the 4th coding level was 42.6\%, approaching the 51–-56\% human‑coder range reported for the same database. Similarly, exposure assessment performance was state-of-the-art. For the ALOHA+-JEM, the evaluation for all individuals resulted in a median PABAK of 0.87 (range: 0.80–0.97), and for exposed individuals, the median PABAK was 0.80 (range: 0.45–0.95). For the SHOCK-JEM, the PABAK was 0.84 for all individuals and 0.67 for exposed-only individuals. These results show OPERAG's framework to be highly effective for accurate, universal AOC.

\subsection{Comparison with current coding tools}
\noindent Differences in classification performance between OPERAG and the existing coding tools AUTONOC, CASCOT, and LabourR may be attributed to the different classification and training frameworks applied. AUTONOC and LabourR were developed using supervised ML techniques trained on other datasets than AsiaLymph, which also utilize classification systems other than ISCO-88. Here, AUTONOC employs a hierarchical ensemble encoder trained on the Canadian National Occupational Classification (NOC)~\citep{garcia2021s}, while LabourR combines keyword lookup with k‑nearest neighbours to classify text into the European Skills, Competences, Qualifications, and Occupations (ESCO) system~\citep{kouretsis2020labourr}. Consequently, since AsiaLymph likely exhibits different textual features than the databases used during training, these models likely overfitted to the distributions and linguistic properties of their training data, resulting in the lower OoD performance on AsiaLymph. Contrastingly, CASCOT relies on a text‑matching algorithm to map inputs to ISCO‑08 and the UK SOC systems, which does not require supervised learning for model development, and is therefore not dependent on training data~\citep{jones2004cascot}. This highlights the importance of approaches that do not require expert-labeled databases to achieve robust OoD coding performance, as they are unable to overfit on the textual properties of specific occupational data used during training. 

Despite both OPERAG and CASCOT being classification frameworks that do not require expert-labeled databases, OPERAG substantially outperforms CASCOT. This gap is likely partly explained by the limitations of CASCOT's approach, which relies on text-matching algorithms that typically compare input text only to predefined patterns or keywords. Such methods do not account for context, miss subtle semantic differences, and struggle to handle incomplete or ambiguous information, which are frequent issues when dealing with short, variable, and often ambiguous occupational data~\citep{hu2019survey}. In contrast, OPERAG leverages LLMs, which possess a deep, generalizable understanding of language. This enables OPERAG to interpret highly variable occupation descriptions as the same occupation, contributing to its higher performance. This was also observed in another RAG-based ISCO-08 classification tool, where job vacancies were classified into ISCO-08, resulting in 58\% accuracy when tested on 36,000 job advertisements~\citep{ge2025tnoauto}. Although this represents a different classification problem due to differences in input length and available context, with job advertisements containing a greater amount of (standardized) terminology than job descriptions, vacancies that should belong to the same occupations may still exhibit different linguistic properties. Consequently, this demonstrates LLMs' ability to handle variable linguistic properties and terminology between occupational texts.

In addition to the difference in classification framework, the performance gap between OPERAG and the other three systems may also be explained by all other systems' using only job titles as input. This approach neglects the discriminative information contained in tasks, employers, or products, which are features that expert coders rely on during classification. Contrastingly, OPERAG integrates all AsiaLymph's free‑text fields into the classification pipeline, enabling this discriminative information to be used during the classification process. This, in combination with the LLM's ability to handle the variable occupational data, likely resulted in the near-human OoD classification performance.

\subsection{Challenge}
\noindent Although OPERAG achieved state‑of‑the‑art performance, it did not perform within AsiaLymph's reported human‑coder range (i.e., 51--56\%)~\citep{ge2021occupational}. Several database‑related factors may explain this gap, namely 1) expert-coder biases, 2) geographical and study-specific assumptions, and 3) AsiaLymph's applied data transformations. 

Occupational coding is inherently variable, as no absolute “gold standard” exists beyond the coding index examples. Consequently, expert coders’ coding habits may differ depending on their training, prior experience, and study context. In AsiaLymph's coding procedure, coders had no prior experience and were trained with feedback~\citep{ge2021occupational}. While this procedure intends to harmonize coding decisions and increase inter‑coder reliability, it may also have introduced systematic biases that reflect learned heuristics. 

Contextual elements such as geography and study population may influence coding outcomes~\citep{morris1990occupational}. For example, in a cohort where most participants are agricultural workers, a job entry labeled as “farmer” may be coded as “agricultural farmer” because this is considered the most likely interpretation. Similarly, although coding to the most detailed level is typically preferred to enhance data quality, occupational epidemiology studies often prioritize pragmatic, more efficient coding choices when coding to a more detailed coding level does not affect study outcomes. For example, in studies assessing risk of electrical shock using the SHOCK‑JEM, where only a 3rd‑level ISCO‑88 code is required, expert coders may assign episodes more quickly to the 3rd level, even though the occupational information might enable coding to the 4th coding level~\citep{huss2013electric}. While not specified for AsiaLymph, its study population stemming solely from Eastern Asia, with its sole application for studying lymphoma, may have introduced geographical and study-specific assumptions in the occupational coding process.

In addition to variability arising from expert‑coder practices and context‑dependent considerations, another important source of variability may stem from the data transformations applied to the AsiaLymph dataset~\citep{ge2021occupational}. The occupational information was originally collected in Chinese and coded into the national SOCS‑ROC system before being translated into English and cross‑walked to ISCO‑88. During this process, linguistic nuances and cultural coding logic may have been lost during translation, resulting in potential semantic misalignments between the English job descriptions and their final ISCO‑88 assignments. Furthermore, although SOCS‑ROC and ISCO‑88 are structurally similar, one‑to‑many crosswalks may introduce ambiguity or a not fully-directly correlating occupational code, as it might not be present in the ISCO-88 classification. Such textual and coding discrepancies may limit OPERAG’s ability to achieve human‑level performance, as the framework is applied only to the English translations and ISCO‑88 codes. 

Although intended to improve inter-coder reliability among expert coders, expert-coder biases, as well as context and study‑dependent practices, performing automatic coding evaluations on databases containing these systematic biases may hamper the accurate reflection of the automatic coding tool's real-world performance. Consequently, OPERAG’s inter‑coder reliability and alignment with the AsiaLymph codes may be improved by formalizing and applying AsiaLymph's context‑specific and study‑dependent rules in OPERAG's framework.

\subsection{Performance evaluation}
\noindent OPERAG’s classification performance varied substantially across major occupational groups, with “Legislators, Senior Officials, and Managers” achieving the lowest accuracy at 13.6\%. This performance is largely attributable to the majority of AsiaLymph's episodes in this group being coded as “Directors and Chief Executives” (i.e., 1210). In practice, these episodes were often classified into unit groups belonging to the sub‑major group “General Managers” (i.e., 1300). Here, the key difference between this sub‑major group and “Corporate Managers” (i.e., 1200) lies in the size of the managed company, which is rarely specified in AsiaLymph's job episodes. Consequently, accurate classification into either group was often impossible without relying on the aforementioned context‑dependent assumptions or rules. For example, AsiaLymph's expert coders may have been trained to code episodes with titles such as “manager” or “boss” into “Directors and Chief Executives” (i.e., 1210) unless other information clearly indicated an alternative code. More likely, however, expert coders used external sources (e.g., company websites) when an employer name was provided, enabling them to infer company size and assign the appropriate code. Without such contextual information, the LLM frequently defaulted to codes corresponding to the generally more common smaller companies, thereby classifying them into unit groups under sub‑major group 13. Moreover, as many of these job episodes contained industry‑specific descriptors (e.g., “factory chief”), the OPERAG's retriever often retrieved industry‑specific minor groups from sub‑major group 13 (e.g., 1312 – General Managers in Manufacturing) due to their semantic similarity, which have no equivalent in sub‑major group 12. Consequently, these or similar entries were most prominent in such episode's top‑30 retrieval results and top‑5 reranked candidates, making it highly unlikely for "1210" to be selected due to LLM’s position bias. A similar effect was observed in the major group “Skilled agricultural and fishery workers” (i.e., "6000"), where the most prevalent category, “Market gardeners and crop growers” (i.e., 6110, accounting for 7.1\% of AsiaLymph episodes), achieved only 3.9\% accuracy. Here, many episodes were instead coded into the more detailed occupational group 6111, reflecting the model’s tendency to assign to more detailed categories even in cases where information does not allow for coding to this level of detail.

Apart from the small Armed Forces group (2.1\% of episodes), the major group "Professionals" obtained the highest classification accuracy. The comparatively strong performance for this group may be explained due to their unit groups being more distinctive, with related occupations often having clearer lexical cues, reducing ambiguity during retrieval and classification~\citep{peterson1999occupational}. Misclassifications that did occur in this major group typically arose when the model deemed the occupational episode to lack sufficient detail to support assignment to less aggregated coding levels. These findings highlight the need to incorporate additional expert‑derived contextual rules into AOC frameworks to improve classification performance.

At each coding level, OPERAG's macro F1 scores were comparable to, or marginally lower than, their respective accuracy. In contrast to supervised learning models, this shows that OPERAG does not exhibit disproportionate prediction of certain classes based on their prevalence in a study population~\citep{yacouby2020probabilistic}. However, the overall macro F1 score (0.39) was still lower than the overall accuracy (42.6\%) on the 4th coding level. This can be attributed to 55 rare ISCO-88 codes achieving F1 scores of zero. As the macro F1 calculates the harmonic mean across all classes and treats each occupational code with equal weight regardless of its frequency, these zero-scoring codes substantially reduced the F1-score. Contrastingly, within individual major groups, the macro F1 scores often equaled or even exceeded the corresponding accuracy. This occurred in groups where the most prevalent unit group (i.e., the 4th coding level) within a major group (i.e., the first coding level) exhibited poor classification performance. For example, in major group "Skilled agricultural and fishery workers" (6), code 6110 accounted for 55.0\% of all episodes but achieved only 3.9\% accuracy. As accuracy is sample-weighted, this majority class heavily influenced the overall major group accuracy, whereas the macro F1 score, being unweighted by sample size, more accurately reflected the typical performance across all occupational codes within the group. These findings demonstrate OPERAG's independence from population-specific occupational distributions, in contrast to conventional supervised machine learning approaches, thereby reinforcing its potential as a universally applicable AOC framework.

OPERAG's exposure assessment performance was substantially higher than its classification performance. This is likely due to the high proportion of occupational codes resulting in “no exposure” within the ALOHA+ and SHOCK-JEMs. Consequently, in case of discordant occupational codes, the resulting exposure assignments may still frequently be the same. Nevertheless, exposure assessment accuracy decreased when analyses were restricted to exposed individuals, for some exposures by as much as 18 percentage points. This highlights an important consideration for real‑world applications, as misclassifying exposed individuals may have a large impact on subsequent epidemiological analyses and can severely impact health outcomes. However, OPERAG's framework can be adapted to study-specific contexts through potential fine-tuning or adaptation of the LLM's classification rules. For example, additional context‑based rules or LLM‑based verification steps could be incorporated to double‑check occupations that have a high likelihood of exposure. Such adaptations further ensure the reliable application of OPERAG's framework in large‑scale epidemiological studies.

\subsection{Limitations}
\noindent While the usage of LLMs enabled the generalizable interpretation of job descriptions and AOC without the need for expert-labeled databases, their highly parameterized architecture precludes full mechanistic explainability due to their reliance on highly nonlinear and distributed computations across millions to billions of parameters~\citep{zhao2024explainability}. Consequently, it is practically infeasible to attribute a specific output to a clearly defined, human-interpretable rule or input feature based on the LLMs' internal mechanisms alone. Contrastingly, glass-box AI models (e.g., symbolic AI) inherently enable mechanistic explainability, as they allow for direct traceability between inputs and outputs through explicit, rule-based logic or structured knowledge representations that can be inspected and verified by human experts. Although the currently employed LLMs provide textual reasoning traces, these are generated post-hoc by the model itself and may not reflect the actual internal decision-making process, leading to potential fabricated or hallucinated justifications for the model's predictions. As such traces could contain plausible-sounding but incorrect reasoning steps, the expert-coder could be misled into choosing an incorrect occupational coding. Especially for less experienced expert-coders, this could lead to misclassifications of occupations and incorrect or biased subsequent exposure assessment. Subsequently, this could lead to severe negative health outcomes or less impactful interventions that promote workers' health. Consequently, it is crucial to take this into account if reasoning traces are shown to experts to ensure accurate occupational coding.

OPERAG's retriever performance is largely determined by the segmentation and embedding of the coding index entries. Although we currently employ structure‑aware chunking, which typically yields strong retrieval performance, alternative chunking strategies could further enhance retrieval efficiency in practice~\citep{stabler2025impact}. For example, we currently include excluded occupations within the chunks, which may introduce semantically confounding information, as these entries frequently contain terminology that has a semantic overlap with closely related occupational groups. Because embedding models are unlikely to distinguish between “included” and “excluded” information, retrieval performance may degrade due to such overlaps. Similarly, while our current embedder was selected due to its state‑of‑the‑art performance across general retrieval benchmarks, these benchmarks do not necessarily reflect the challenges of short‑text occupational classification~\citep{kenter2015short}. In settings where input texts are particularly short and ambiguous, embedders trained on corpora that more closely match the linguistic characteristics of occupational data could outperform the currently used general‑purpose embedder. Consequently, systematically evaluating alternative chunking strategies and embedders across occupational datasets may improve OPERAG's retrieval performance.

As cross-encoders are typically optimized for broader retrieval tasks and may not adequately disambiguate between closely related occupational codes, we adopted a reasoning‑LLM as reranker~\citep{dejean2024thorough}. The reasoning‑based reranker produced improvements, increasing top‑5 retrieval accuracy and thereby improving classification performance. However, the improvement was limited (4.3 percentage points), partly because in 12.2\% of cases the gold standard code was absent from the initial top‑30 retrieval results. This absence may reflect limitations in semantic similarity search when occupational information lacked semantic overlap with index entries. For example, the ISCO‑88 group, “Administrative Associate Professionals not classified elsewhere”, comprising 4.9\% of AsiaLymph's job episodes, was often not retrieved by the RAG-components. This is likely due to the corresponding occupational information often being described more generally rather than being semantically similar (e.g., "Director"). This resulted in the occupational information likely being embedded further away from the coding index entry in the embedding space, disabling effective retrieval. Consequently, alternative embedding strategies, or text augmentation strategies using (e.g., applying thesauruses to find synonyms), could be employed to improve retrieval performance.

We employed Qwen3‑14B‑AWQ as the classifier component, as it provides strong reasoning capabilities relative to its parameter size while remaining suitable for local deployment on consumer‑grade hardware~\citep{yang2025qwen3}. This design choice is especially important given that occupational data are often proprietary, frequently precluding inference on third‑party infrastructure. Regardless, several limitations remain. First, reasoning LLMs with a larger number of parameters could potentially yield further performance improvements due to their enhanced capacity for complex reasoning~\citep{guo2025deepseek}. Second, the model used in this study is a quantized version of the original model, meaning that its weights were compressed from full precision (i.e., 16-bit BF16) into lower precision formats (i.e., 4-bit) to reduce memory requirements and accelerate inference~\citep{egashira2024exploiting}. While quantization is essential to enable efficient local deployment on consumer‑grade hardware, it often degrades performance, especially for smaller models~\citep{marchisio2024does}. However, quantized model versions often still outperform other models utilizing all their parameters, while both employ the same amount. Third, alternative model architectures with pretraining corpora more closely aligned to occupational texts, or models fine‑tuned specifically for this domain, may outperform the general‑purpose LLM employed here. Consequently, although our framework currently prioritizes generalizability, adaptability to different classification systems, and local deployability, an evaluation of alternative architectures and model sizes, particularly those trained or adapted on representative and domain‑specific occupational datasets, may further improve OoD automatic occupational classification performance.

Currently, OPERAG was evaluated solely using the AsiaLymph dataset using only the ISCO-88 classification. Although the framework was explicitly designed to be independent of expert-labeled data while being capable of adapting to diverse classification systems, its current validation does not guarantee similar performance across other datasets or occupational classifications. As occupational data can differ substantially in linguistic characteristics and apply other classification systems with highly differing taxonomies, model performance likely varies. Consequently, to enable a more comprehensive understanding of OPERAG framework’s generalizability, robustness, and sensitivity to dataset- or classification‑specific characteristics, it should be evaluated using multiple occupational datasets coded into different occupational classification systems.

\subsection{Future Work}
\noindent OPERAG's framework can be extended to improve its classification performance in both database-dependent and independent settings, namely by: 1) allowing for confidence scores to be obtained for each prediction, 2) integrating active learning, and 3) using current LLMs' agentic features to enable automatic retrieval of external web-based information based on the user's query.

In most (semi-)AOC tools, confidence scores are provided alongside each prediction~\citep{russ2016computer, langezaal2023artificial, jones2004cascot}. These scores estimate the likelihood that a given code is correct. When there is strong calibration between the confidence score and the actual probability of correctness (e.g., a prediction with a 95\% confidence score is correct in 95\% of cases), expert coders can establish thresholds above which predictions are accepted automatically. This strategy is commonly applied in semi‑automatic or decision‑supported occupational coding. In such scenarios, codes that exceed the pre‑defined confidence threshold are accepted without further review, while the remaining cases are presented to the expert coder for manual verification and potential correction. This strategy combines the efficiency of automatic coding with the reliability of manual coding. However, due to the autoregressive nature of LLMs, they are inherently unable to provide reliable confidence scores for their predicted classes~\citep{geng2023survey}. Consequently, without the use of a calibration set, two suboptimal alternatives remain: (1) approximating confidence based on token‑level logit distributions, or (2) prompting the model to self‑report its confidence. Both approaches have been shown to produce poorly calibrated estimates when compared to the confidence scores generated by conventional ML classifiers. To address this limitation, calibration mechanisms can be introduced in which model outputs are calibrated using a subset of expert-annotated occupational codes and confidence scores, thereby establishing reliable task‑specific confidence thresholds. Regardless, because OPERAG was designed as a framework that does not rely on expert-labeled databases, the generation and use of such a calibration set was not feasible within the scope of this study. In real‑world deployments, however, expert coders could annotate a limited initial subset of job descriptions with associated confidence scores, enabling the creation of a calibration set. This would allow OPERAG to generate reliable confidence estimates, thereby supporting its integration into semi‑automatic coding workflows. Such an extension would enhance both the efficiency and reliability of occupational coding in large‑scale epidemiological studies and related applications.

While the current framework enables accurate AOC without requiring expert-labeled databases, its performance has not yet reached human-level accuracy. This limitation can partly be attributed to the framework’s inability to adapt to expert-coded occupations, which implicitly reflect the aforementioned expert-coder bias, as well as context and study-dependent rules required to achieve human performance. Extending OPERAG with mechanisms that iteratively improve expert-alignment based on feedback could therefore enable more stable and reliable occupational classification~\citep{lu2023human}. Within the current framework, such adaptation could be achieved either through periodic fine‑tuning of the classification component or through incremental updates to the retrieval component with corrected entries. Although fine‑tuning reasoning models is comparatively resource‑intensive, as it requires the construction of a reasoning database containing a high number of complete reasoning traces combined with correct outcomes, this has been shown to achieve state-of-the-art classification performance~\citep{guo2025deepseek}. Potentially, certain aspects of the reasoning-trace generation process can be partially automated to reduce the manual burden of manually constructing these~\citep{whitehouse2023llm}. Alternatively, a lightweight adaptation mechanism could be employed, in which a supplementary retrieval system identifies previously coded episodes that are semantically similar to the input and integrates these, together with the corrected code, into the prompt as context. Such a mechanism may further enhance the robustness and stability of AOC.

Currently, expert coders often employ context-specific information beyond the provided raw free-text occupational information to support their coding decisions. For example, while ISCO-88 groups "Directors and Chief Executives" and "General Managers" share highly similar tasks, the main differentiating aspect is the size of the company that the person manages~\citep{peterson1999occupational}. When provided with a company name, expert coders may consult external sources (e.g., internet searches) to determine the company's industry, primary products, or organizational size, thereby enabling more definitive classification in either group. In contrast, the present OPERAG framework may treat the company name as plain contextual input if this company's information is not present in the LLM's pre-training corpus. If no additional information is supplied within the occupational description itself, the model lacks access to further contextual knowledge, limiting its ability to fully mimic this expert‑coder decision process and thereby constraining performance relative to human coders. However, recent advances in LLMs have introduced agentic capabilities, such as web‑browsing, which allow models to dynamically query external resources when additional information is required based on the input~\citep{plaat2025agentic}. By integrating such functionality, the framework could autonomously retrieve and leverage relevant contextual occupational data, further improving its classification performance.

\subsection{Conclusions}
\noindent We show that OPERAG is a state-of-the-art framework for database and classification-independent occupational coding, outperforming current coding tools on OoD occupational classification performance while approaching human performance. OPERAG's classification performance may be improved by refining retrieval strategies, adapting to study and context‑specific coding rules, and extending the framework with confidence calibration, active learning, and agentic capabilities for external information retrieval. Additionally, OPERAG's evaluation should be extended to include additional occupational datasets and classification systems, thereby improving insight into the influence of embedder choice, model architecture, and coding‑system variability on classification performance. Consequently, the OPERAG framework may unlock current and future occupational data potential in finding crucial exposure response associations, leading to new interventions that improve workers' health.

\subsection*{Data availability}
\noindent The AsiaLymph~\citep{friesen2016evaluation} dataset is not publicly available. Access to this dataset should be requested from the authors of the original study. The ALOHA+-JEM~\citep{matheson2005biological} is not publicly available and should be requested from the authors of the original study. The SHOCK-JEM~\citep{huss2013electric} is publicly available, where the matrices for their use in the context of epidemiological studies can be found in the original study.

\subsection*{Declaration of generative AI and AI-assisted technologies in the manuscript preparation process}
\noindent During the preparation of this work, the authors used Gemini 2.5 and 3 Pro for textual improvement of the manuscript by checking its spelling, grammar, and textual flow. After using this tool/service, the authors reviewed and edited the content as needed and takes full responsibility for the content of the published article.

\appendix
\section{Example chunk of ISCO-88 occupational group in OPERAG's knowledge base}\label{app:chunk}
\noindent ``\textbf{ISCO-88 Code 6111}: Field crop and vegetable growers\\
\noindent\textbf{Definition}\\
\noindent Field crop and vegetable growers plan and carry out the necessary operations to grow and harvest various types of field crops such as wheat and other cereals, rice, beetroot, sugar-cane, ground-nuts, tobacco, reeds or other field crops and potatoes, cabbages or other field vegetables, for sale or delivery on a regular basis to wholesale buyers, marketing organisations or at markets.\\
\noindent\textbf{Tasks}
\begin{itemize}
    \item Determining kinds and amounts of crops to be grown
    \item Purchasing seeds, fertiliser and other supplies
    \item Renting or investing in land and land improvements, buildings, working animals, equipment and machinery
    \item Performing farm operations such as land preparation, sowing, planting, cultivating and harvesting crops and field vegetables
    \item Storing and carrying out some basic processing of their produce
    \item Tending working animals and maintaining farm buildings, machinery and equipment
    \item Delivering or marketing farm products
    \item Performing related tasks
    \item Supervising other workers
\end{itemize}
\noindent \textbf{Included Occupations}
\begin{itemize}
    \item Farm worker, skilled/field crops
    \item Farmer, cereal
    \item Farmer, cotton
    \item Farmer, field crop
    \item Farmer, rice
    \item Farmer, sugar-cane
\end{itemize}
\noindent \textbf{Occupations Classified Elsewhere}
\begin{itemize}
    \item Department manager, production and operations/agriculture (1221)
    \item General manager, agriculture (1311)
    \item Labourer, farm (9211)
    \item Managing director, enterprise/agriculture (1210)
    \item Operator, agricultural machinery (8331)
\end{itemize}

\section{OPERAG's reranking prompt}\label{app:rerank}
\noindent\textbf{System prompt:} \\ ``You are an expert coder specializing ONLY in the International Standard Classification of Occupations version 88 (ISCO-88).''
\\\\
\textbf{User prompt:}\\
``\noindent\textbf{Coding index entries:}\\
\noindent[List of 30 coding index entries as provided in Appendix A.]\\
\noindent\textbf{Objective:} Based ONLY on the input data and the provided ISCO-88 coding index context, propose 5 distinct ISCO-88 codes that are most likely for the given occupation.
\noindent\textbf{Classification Rules:}
\begin{enumerate}
    \item Review all provided input data (Occupation Title, Primary Tasks, Employer, Product/Service) to understand the job's context, duties, and required skill level.
    \item Contextualize with Employer Information: Use `Employer' and `Employer's Main Product/Service' to establish the work's operational context (e.g., scale, industry, technology). This context must guide the interpretation of `Occupation Title' and `Primary Tasks', aiding differentiation between ISCO codes that vary by their implied operational environment or skill specialization.
    \item Evaluate applicability hierarchically from general (e.g., coding level 1) to specific (e.g., coding level 4). A more specific code (e.g., level 4) can only be selected if the input data provides concrete details that distinguish it from other codes at the same specific level (i.e., its sibling codes). If the provided information is too generic to differentiate between multiple sibling codes, the more general parent code must be chosen. For instance, consider an ``Engine Technician'' whose task is ``repairing engines.'' This generally fits under ``Vehicle Mechanics.'' However, without information specifying the type of vehicle (car, motorcycle, heavy truck, etc.), one cannot definitively choose between the more specific codes for ``Car Mechanic'' or ``Heavy Truck Mechanic.'' In this scenario, the more general parent code for ``Vehicle Mechanics'' is the correct choice. Therefore, only select a specific code if the combined input data (Occupation Title, Primary Tasks, Employer, Product/Service) provides details that uniquely and exclusively point to that code. Do not make any assumptions that are not explicitly substantiated by the input.
    \item \textbf{CRITICAL:} Base your decision \textit{solely} on the \textbf{ISCO-88 coding index context} provided above. Do \textit{not} use external knowledge of ISCO-88.
    \item For each \textbf{ISCO-88 coding index context}, Analyze \textit{why} it fits or \textit{why} it does not fit the \textbf{Input Data}, referencing specific details from both the input and the context.
    \item Output \textit{only} the final code in the specified format.
\end{enumerate}
\textbf{Input Data (Current Job to Classify):}
\begin{itemize}
    \item \textbf{Occupation Title:} Farming
    \item \textbf{Primary Tasks:} Planting
    \item \textbf{Employer:} Going up to the mountains and down to the countryside
    \item \textbf{Employer's Main Product/Service:} Planting
\end{itemize}
\noindent \textbf{Output Format:}
Provide your top 5 most likely 4-digit ISCO-88 codes. List them comma-separated after ``Codes:". Include a brief reasoning for your selection of these codes.
Example: Codes: 1234, 5678, 2345
Reasoning: [Briefly explain why these codes are plausible candidates based on the input and context.]\\
\noindent \textbf{Begin Classification}''

\section{OPERAG's classification prompt}\label{app:classify}
\noindent\textit{\textbf{System prompt:}}\\ ``You are an expert coder specializing ONLY in the International Standard Classification of Occupations version 88 (ISCO-88)."
\\\\
\textit{\textbf{User prompt:}} \\
``\noindent\textbf{Coding index entries:}\\
\noindent[List of 5 coding index entries as provided in Appendix A.]\\
\noindent\textbf{Objective:} Classify the provided occupation into the most appropriate ISCO-88 code based ONLY on the input data and the provided ISCO-88 coding index context. \\
\noindent\textbf{Classification Rules:}
\begin{enumerate}
    \item Review all provided input data (Occupation Title, Primary Tasks, Employer, Product/Service) to understand the job's context, duties, and required skill level.
    \item Contextualize with Employer Information: Use `Employer' and `Employer's Main Product/Service' to establish the work's operational context (e.g., scale, industry, technology). This context must guide the interpretation of `Occupation Title' and `Primary Tasks', aiding differentiation between ISCO codes that vary by their implied operational environment or skill specialization.
    \item Evaluate applicability hierarchically from general (e.g., coding level 1) to specific (e.g., coding level 4). A more specific code (e.g., level 4) can only be selected if the input data provides concrete details that distinguish it from other codes at the same specific level (i.e., its sibling codes). If the provided information is too generic to differentiate between multiple sibling codes, the more general parent code must be chosen. For instance, consider an ``Engine Technician'' whose task is ``repairing engines.'' This generally fits under ``Vehicle Mechanics.'' However, without information specifying the type of vehicle (car, motorcycle, heavy truck, etc.), one cannot definitively choose between the more specific codes for ``Car Mechanic'' or ``Heavy Truck Mechanic.'' In this scenario, the more general parent code for ``Vehicle Mechanics'' is the correct choice. Therefore, only select a specific code if the combined input data (Occupation Title, Primary Tasks, Employer, Product/Service) provides details that uniquely and exclusively point to that code. Do not make any assumptions that are not explicitly substantiated by the input.
    \item \textbf{CRITICAL:} Base your decision \textit{solely} on the \textbf{ISCO-88 coding index context} provided above. Do \textit{not} use external knowledge of ISCO-88.
    \item For each \textbf{ISCO-88 coding index context}, Analyze \textit{why} it fits or \textit{why} it does not fit the \textbf{Input Data}, referencing specific details from both the input and the context.
    \item Output \textit{only} the final code in the specified format.
\end{enumerate}
\textbf{Input Data (Current Job to Classify):}
\begin{itemize}
    \item \textbf{Occupation Title:} Farming
    \item \textbf{Primary Tasks:} Planting
    \item \textbf{Employer:} Going up to the mountains and down to the countryside
    \item \textbf{Employer's Main Product/Service:} Planting
\end{itemize}
\noindent \textbf{Output Format:}
Provide only the final classification in the following format:\\
\textbf{Code:} [ISCO-88 code]\\
\textbf{Reasoning:} [Explain step-by-step why you chose this code, referencing comparisons between Input Data and context.]\\
\noindent \textbf{Begin Classification}"

\bibliographystyle{elsarticle-harv} 
\bibliography{sn-bibliography}

\end{document}